\documentclass{article}
\usepackage{iclr2027_conference,times}
\usepackage[OT1]{fontenc}

\usepackage{amsmath,amsfonts,bm}

\def\eqref#1{equation~\ref{#1}}

\def\1{\bm{1}}

\DeclareMathAlphabet{\mathsfit}{\encodingdefault}{\sfdefault}{m}{sl}
\SetMathAlphabet{\mathsfit}{bold}{\encodingdefault}{\sfdefault}{bx}{n}

\usepackage{url}
\usepackage{graphicx}
\usepackage{booktabs}
\usepackage{amsmath}
\usepackage{amsfonts}
\usepackage{amssymb}
\usepackage{amsthm}
\usepackage{xcolor}
\usepackage{algorithm}
\usepackage{algpseudocode}
\usepackage{pifont}      
\usepackage{listings}    
\usepackage{hyperref}
\renewcommand{\eqref}[1]{\textup{(\ref{#1})}}

\definecolor{retraqInk}{HTML}{22262A}    
\definecolor{retraqBlue}{HTML}{2B6CB0}   
\definecolor{retraqGreen}{HTML}{1F8A70}  
\definecolor{retraqGray}{HTML}{6B7480}   

\newcommand{\stepA}{\mbox{\textcolor{retraqInk}{\ding{202}}}}
\newcommand{\stepB}{\mbox{\textcolor{retraqBlue}{\ding{203}}}}
\newcommand{\stepC}{\mbox{\textcolor{retraqGreen}{\ding{204}}}}

\lstdefinestyle{retraqcode}{
  language=Python,
  basicstyle=\fontsize{7.6pt}{8.7pt}\ttfamily\selectfont,
  columns=fullflexible,
  keepspaces=true,
  keywordstyle=\bfseries,
  commentstyle=\color{retraqGray},
  showstringspaces=false,
  breaklines=false,
  escapeinside={(*@}{@*)},
  aboveskip=0pt,
  belowskip=0pt,
  xleftmargin=2pt,
}

\newtheorem{proposition}{Proposition}

\title{RetraQ: Manifold-Constrained Critic\\
Guidance for Flow Policies in RL}

\author{%
Xuanzhuo Liu$^{1,2}$\thanks{Work done during an internship at AgiBot.}\hspace{0.5em},
Wu Chen$^{2}$, Ruixiang Wu$^{3}$, Shuai Yuan$^{2,4}$\footnotemark[1] \\
\textbf{Chuankang Li}$^{2}$\thanks{Corresponding authors: \texttt{lichuankang@agibot.com}, \texttt{litongxin@cuhk.edu.cn}.}\hspace{0.5em},
\textbf{Tongxin Li}$^{3}$\footnotemark[2]\hspace{0.5em},
\textbf{Maoqing Yao}$^{2}$, \textbf{Guanghui Ren}$^{2}$ \\
$^{1}$Stanford University\quad
$^{2}$AgiBot\quad
$^{3}$CUHK-Shenzhen\quad
$^{4}$Southeast University
}

\iclrfinalcopy
\AddToHook{cmd/maketitle/after}{%
  \fancyhead{}%
  \renewcommand{\headrulewidth}{0pt}%
}

\begin{document}

\maketitle
\begin{abstract}
A learned critic can improve a pretrained flow-matching policy by steering
sampled actions toward higher predicted value. However, even a critic that 
fits the behavior data well can provide misleading action gradients.
Gradient components pointing away from the data-supported region are
weakly constrained by behavior data and can lead guidance to exploit
critic extrapolation errors. 
We introduce  \textbf{RetraQ} (\textbf{Retra}ction-based \textbf{Q}-guidance),
which uses geometry encoded by the pretrained flow to constrain both
the \emph{direction} and the \emph{destination} of critic guidance.
A single velocity evaluation of the frozen flow defines a denoising map.
RetraQ filters the critic gradient with the map's Jacobian and applies
the map itself to correct the updated action.
Our analysis characterizes when these operations approximate tangent
filtering and retraction onto the behavior manifold.
The same update guides actions at test time and provides targets for
policy distillation and online reinforcement learning.
In controlled studies, RetraQ improves true return while keeping refined
actions close to the data-supported region, and its distillation targets
remain more stable than those of raw critic ascent.
In offline-to-online training on LIBERO and OGBench, RetraQ achieves
higher mean final success than the evaluated critic-guidance baselines.
On a real-world plug-insertion task, test-time RetraQ improves the
success rate of a behavior-cloned $\pi_{0.5}$ policy from
$55.0\%$ to $76.7\%$. Online training further increases success to $98.3\%$
in only 30 minutes and $100\%$ within one hour.

{\centering
\textbf{Project page:}
\url{https://xuanzhuoliu.github.io/RetraQ/}\par}
\end{abstract}

\section{Introduction}
\label{sec:intro}

Generative policies can model complex, multimodal action distributions that arise in robot manipulation~\citep{florence2022ibc,shafiullah2022bet}.
Flow-matching policies, in particular, learn a velocity field that transports noise into actions and can generate action chunks with only a few deterministic integration steps~\citep{lipman2023flow,liu2023rectified}.
Their efficient sampling and simple regression objective make them well suited to behavioral cloning~(BC).
Recent vision-language-action policies have adopted flow matching as their core action-generation mechanism~\citep{black2024pi0,intelligence2025pi05,bjorck2025gr00t}.

However, behavior cloning does not directly optimize task outcomes and can 
inherit suboptimal behavior from its training data.
In precise manipulation, a policy may complete most of a task yet still fail at
a critical stage, such as aligning a plug before insertion.
Reinforcement learning~(RL) provides a natural way to improve such behaviors
using reward feedback~\citep{luo2024hilserl,intelligence2025pistar06}.
For a pretrained flow policy, this feedback can be incorporated through a
learned critic: following its action gradient
$\nabla_{\mathbf a}Q(\mathbf s,\mathbf a)$ can steer sampled actions toward
higher predicted return~\citep{silver2014dpg,
wang2023diffusionql,psenka2024qsm,park2025fql}.

\begin{figure}[t]
\centering
\includegraphics[width=\textwidth]{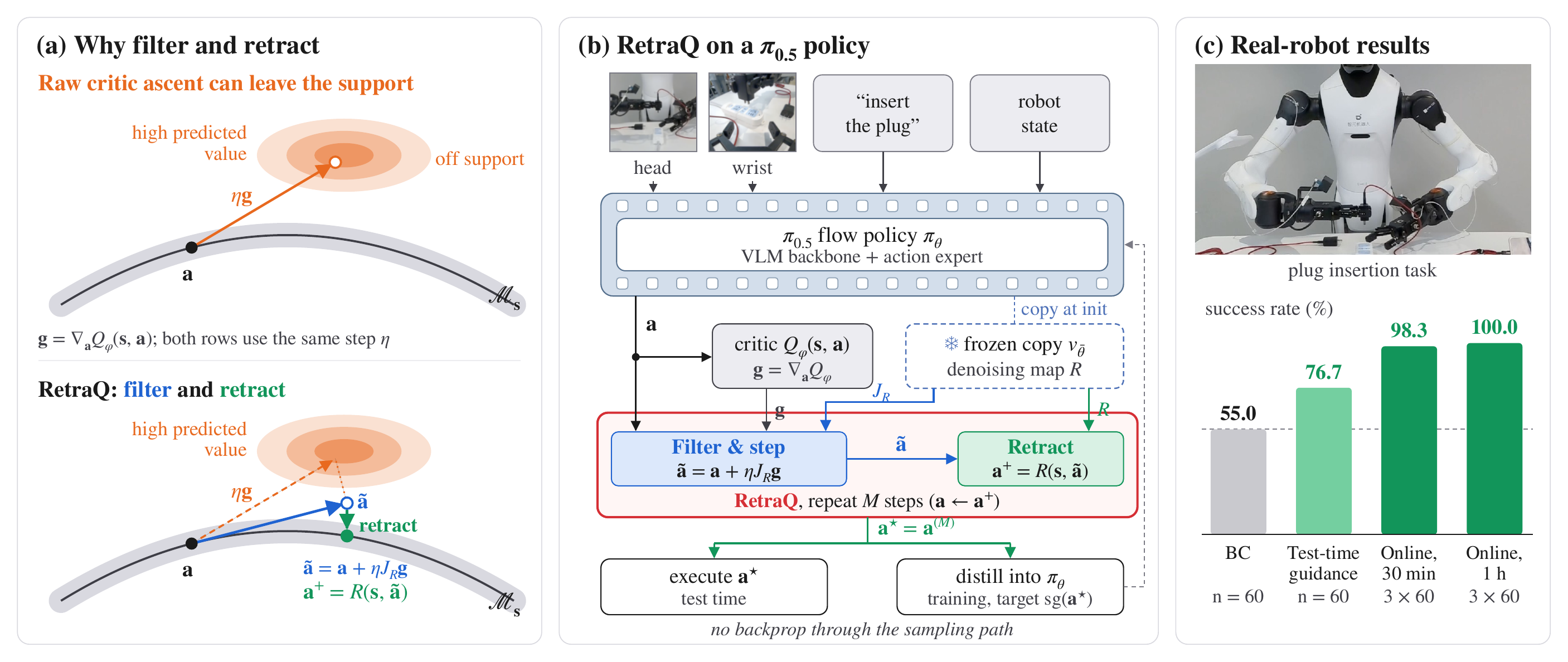}
\caption{
\textbf{RetraQ at a glance.}
\textbf{Left:} raw critic ascent can move actions away from the behavior
manifold $\mathcal M_{\mathbf s}$ by following normal gradient components.
RetraQ uses the Jacobian $J_R$ of a frozen flow's denoising map $R$ to filter
the gradient and uses $R$ to correct the updated action, constraining the
update direction and destination.
\textbf{Right:} for a $\pi_{0.5}$ policy, a frozen copy of the flow defines
$R$.
The refined action $\mathbf a^{+}$ is executed at test time or used as a
stop-gradient training target, without backpropagating through the policy's
sampling trajectory.
}
\label{fig:teaser}
\end{figure}

The difficulty is that a critic can predict returns well for actions in the training data, 
yet its gradients may still provide unreliable directions for policy improvement.~\citep{mage}.
Consider the idealized case in which behavior actions at a fixed context lie
on a smooth, lower-dimensional manifold.
Two differentiable critics that agree on a local patch of the manifold must
have the same tangential derivatives, while their normal derivatives can
differ arbitrarily.
Thus, even exact critic values on the data support leave the normal component
of the action gradient undetermined.
Following such a component can move an action away from the data-supported
region, where a high predicted value may reflect critic extrapolation error
rather than true improvement~\citep{fujimoto2019offpolicy}.
Filtering the direction alone is not sufficient: a finite step along a tangent
direction can still leave a curved manifold.
Reliable critic guidance therefore requires constraining both the update
\emph{direction} and its \emph{destination}.

We introduce \textbf{RetraQ}, which
approximates both operations using geometry already encoded by a pretrained
flow policy.
At a chosen flow time, a frozen flow defines a denoising map from a noisy
action to an estimate of its clean counterpart.
RetraQ uses the Jacobian of this map to filter the critic gradient and applies
the map after the ascent step to retract the updated action toward the
data-supported manifold (Figure~\ref{fig:teaser}).
Theoretically, we show that the denoising map approximates
a retraction while its Jacobian approximates tangent-space projection,
and we bound how errors of a learned velocity field enter both.
Unlike denoising-based Riemannian optimization of a known
objective~\citep{landingscore}, RetraQ follows a learned critic with
unidentified normal gradients and obtains both operators from a flow policy.
RetraQ requires no additional geometry model, and the filtering step is
implemented with a Jacobian--vector product rather than backpropagation
through the full sampling trajectory.
The same refinement update applies to test-time guidance, offline and online policy distillation.

We evaluate RetraQ in controlled problems, simulation benchmarks, and
real-world plug insertion.
In a controlled 16-dimensional problem with a known manifold, the learned
flow's Jacobian attenuates normal gradient components, and RetraQ keeps
refined actions closest to the manifold among the compared critic-guided
updates.
In these controlled studies, RetraQ also attains higher true return and more
stable distillation targets than raw critic ascent.
In offline-to-online training on LIBERO and OGBench, RetraQ attains
higher mean final success than the evaluated critic-guidance baselines.
On real-world plug insertion, test-time RetraQ raises the success rate of
a behavior-cloned $\pi_{0.5}$ policy~\citep{intelligence2025pi05} from
$55.0\%$ to $76.7\%$, without
updating the policy parameters.
With online training, success reaches $98.3\%$ within 30 minutes and
$100\%$ after one hour.

Our contributions are:
\begin{itemize}
\item \textbf{Geometric analysis of critic-guidance.}
We analyze failure modes of critic guidance and establish conditions where flow denoising approximates tangent projection and retraction.

\item \textbf{Gradient filtering with flow retraction.}
We introduce RetraQ, which uses a pretrained flow's Jacobian to filter critic
gradients and its denoising map to retract updated actions.

\item \textbf{Sim and real-robot validation.}
We validate the geometric mechanism and distillation stability in controlled
studies, and demonstrate policy improvement in simulation and on a real-world contact-rich manipulation task.
\end{itemize}

Section~\ref{sec:related}  discusses connections to denoising-based
manifold optimization and related flow-policy improvement methods, and
Appendix~\ref{app:related-work} extends the comparison.
\section{Preliminaries}
\label{sec:prelim}

\textbf{Flow-matching policies.}
Let $\mathbf s$ denote the context, including the observation, robot state,
and task instruction, and let $\mathbf a\in\mathbb R^D$ be an action chunk
that concatenates a sequence of control commands.
Let $\mu(\cdot\mid\mathbf s)$ denote the behavior action distribution.
During training, flow matching samples
$\mathbf a\sim\mu(\cdot\mid\mathbf s)$ and independent noise
$\mathbf z\sim\mathcal N(0,I_D)$, and forms
$ 
    \mathbf x_t=t\,\mathbf a+(1-t)\,\mathbf z, t\in[0,1].
    \label{eq:fm-interpolation}
$
The velocity field $v_\theta(\mathbf s,\mathbf x,t)$ is trained by
squared-error regression to $\mathbf a-\mathbf z$~\citep{lipman2023flow,liu2023rectified}.
The policy generates actions by numerically integrating
$\mathrm d\mathbf x/\mathrm dt=v_\theta(\mathbf s,\mathbf x,t)$
from $\mathbf x(0)=\mathbf z$ to $t=1$.
For $t<1$, a single velocity evaluation defines the clean-action estimate
\begin{equation}
    D_t^\theta(\mathbf s,\mathbf x)
    :=\mathbf x+(1-t)\,v_\theta(\mathbf s,\mathbf x,t).
    \label{eq:clean-action-map}
\end{equation}
Since
$\mathbf x_t+(1-t)(\mathbf a-\mathbf z)=\mathbf a$,
under the population-optimal velocity
$
v^\star(\mathbf s,\mathbf x,t)
=
\mathbb E[\mathbf a-\mathbf z
\mid \mathbf x_t=\mathbf x,\mathbf s],
$
this map equals the posterior mean,
$
D_t^\star(\mathbf s,\mathbf x)
=
\mathbb E[\mathbf a
\mid \mathbf x_t=\mathbf x,\mathbf s].
$

\textbf{Behavior manifold.}
For each fixed context $\mathbf s$, we locally model the data-supported
action set as a smooth $k$-dimensional embedded manifold
$\mathcal M_{\mathbf s}\subset\mathbb R^D$, with $k<D$.
When $\mathbf s$ is fixed, we write $\mathcal M=\mathcal M_{\mathbf s}$.
At $\mathbf a\in\mathcal M$, the tangent space
$T_{\mathbf a}\mathcal M$ describes first-order variations along the
manifold, while the normal space $N_{\mathbf a}\mathcal M$ is its
orthogonal complement.
Let $P_T(\mathbf a)$ and
$P_N(\mathbf a)=I_D-P_T(\mathbf a)$ denote the corresponding orthogonal
projectors.
We use the Euclidean metric in the flow's normalized action coordinates.
Let $\Pi_{\mathbf s}(\mathbf x)$ denote the point on
$\mathcal M_{\mathbf s}$ closest to $\mathbf x$ in Euclidean distance.
We use this nearest-point projection locally where the closest
point is unique.

\textbf{Critic-guided refinement.}
Let $Q(\mathbf s,\mathbf a)$ be a learned estimate of expected return,
differentiable in the action.
Holding $\mathbf s$ fixed, we write
$Q(\mathbf a):=Q(\mathbf s,\mathbf a)$ and
$\mathbf g(\mathbf a):=\nabla_{\mathbf a}Q(\mathbf a)$.
We study refinements of an action sampled from the pretrained flow policy:
\begin{equation}
    \mathbf a^+
    =\mathcal C\!\left(
        \mathbf a+\eta F(\mathbf a)\mathbf g(\mathbf a)
      \right),
    \qquad \eta>0,
    \label{eq:refinement-template}
\end{equation}
where $F(\mathbf a)\in\mathbb R^{D\times D}$ transforms the
critic-gradient direction and $\mathcal C$ corrects the resulting endpoint.
Their dependence on $\mathbf s$ is left implicit.
Raw critic ascent corresponds to $F=I_D$ and
$\mathcal C=\mathrm{id}$.
The refined action can be executed directly or used as a regression target
for the flow policy.

Appendix~\ref{app:related-work} addresses existing
critic-guided flow-policy improvement methods.
Section~\ref{sec:identifiability} motivates tangent filtering and retraction,
and Section~\ref{sec:method} approximates both operations with the pretrained flow.

\section{Geometric Constraints on Critic Guidance}
\label{sec:identifiability}

\begin{figure}[t]
    \centering
    \includegraphics[width=0.95\linewidth]{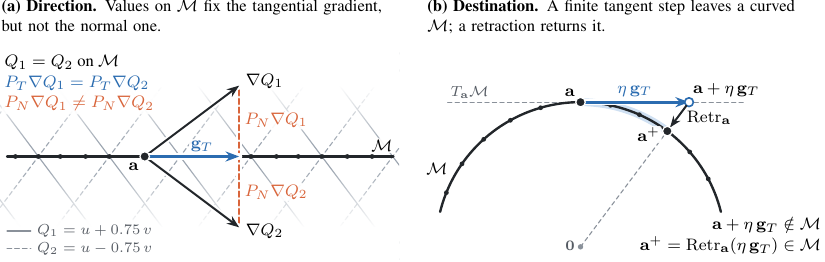}
    \caption{
    \textbf{Constraints on the update direction and destination.}
    \textbf{(a)} The critics $Q_1(u,v)=u+0.75v$ and
    $Q_2(u,v)=u-0.75v$ agree on the behavior manifold
    $\mathcal M=\{(u,0):u\in\mathbb R\}$.
    Their tangential gradients agree, while their normal components have
    opposite signs.
    \textbf{(b)} A finite tangent step from $\mathbf a$ leaves the circular
    manifold. Projecting the endpoint back onto the circle yields
    $\mathbf a^+$ and locally defines a retraction.
    }
    \label{fig:critic-identifiability}
\end{figure}

Fix the context $\mathbf s$, write $\mathcal M=\mathcal M_{\mathbf s}$ and
$\Pi=\Pi_{\mathbf s}$, and consider an action $\mathbf a\in\mathcal M$.
Suppose two continuously differentiable critics agree on a local patch of
$\mathcal M$ around $\mathbf a$.
Because their values change identically along every curve within this patch,
their tangential gradients at $\mathbf a$ must agree.
Their normal gradients, however, are unconstrained by these on-manifold
values and can differ arbitrarily
(Figure~\ref{fig:critic-identifiability}a).
Thus, even exact critic values on the behavior manifold do not determine the
full action gradient.
Appendix~\ref{app:geometric-refinement} formalizes this observation and
extends it to data concentrated in a thin neighborhood of the manifold.

\textbf{Direction: filter the critic gradient.}
Let $\mathbf g=\nabla_{\mathbf a}Q(\mathbf a)$ and decompose
\[
    \mathbf g=\mathbf g_T+\mathbf g_N,
    \qquad
    \mathbf g_T=P_T(\mathbf a)\mathbf g,\quad
    \mathbf g_N=P_N(\mathbf a)\mathbf g.
\]
The tangential component $\mathbf g_T$ is determined by how the critic varies
along the behavior manifold, whereas $\mathbf g_N$ depends on how the critic
extends away from it.
Following $\mathbf g_N$ can move the action into regions weakly supported by
the behavior data, where high predicted value may reflect critic
extrapolation error rather than true improvement
~\citep{fujimoto2019offpolicy}.
We therefore retain only the tangential component,
corresponding to $F(\mathbf a)=P_T(\mathbf a)$ in
Eq.~\eqref{eq:refinement-template}.

\textbf{Destination: correct the endpoint.}
Filtering the direction alone is insufficient.
Because the tangent space is only a local linear approximation, a finite step
$\mathbf a+\eta\mathbf g_T$ can leave a curved manifold even when
$\mathbf g_T$ is exactly tangent
(Figure~\ref{fig:critic-identifiability}b).
For a sufficiently small tangent displacement
$\boldsymbol\xi\in T_{\mathbf a}\mathcal M$, nearest-point projection defines
the local retraction
\[
    \operatorname{Retr}_{\mathbf a}(\boldsymbol\xi)
    :=\Pi(\mathbf a+\boldsymbol\xi),
\]
which returns the stepped action to $\mathcal M$ while preserving the
tangent displacement to first order~\citep{boumal}.
Choosing $F(\mathbf a)=P_T(\mathbf a)$ and $\mathcal C=\Pi$ therefore gives
the geometric update
\begin{equation}
    \mathbf a^+
    =\Pi\bigl(\mathbf a+\eta\mathbf g_T\bigr)
    =\operatorname{Retr}_{\mathbf a}(\eta\mathbf g_T).
    \label{eq:ideal-manifold-update}
\end{equation}
Because $\mathbf g_T$ is determined entirely by the critic values on the
local manifold patch, this update is unchanged by arbitrary differences in
normal critic derivatives.

\textbf{Local improvement guarantee.}
Let $Q^\pi$ denote the true action-value function when the refined action
chunk is followed by a fixed policy $\pi$.
If the learned critic $Q$ agrees with $Q^\pi$ on the local manifold patch,
then their tangential gradients also agree there.
Consequently, under the smoothness conditions of
Proposition~\ref{prop:local-geometric-ascent}, the ideal update in
Eq.~\eqref{eq:ideal-manifold-update} strictly increases $Q^\pi$ for all
sufficiently small positive step sizes whenever
$\mathbf g_T\neq\mathbf 0$.
Formal statements and proofs are given in
Appendix~\ref{app:geometric-refinement}.

In practice, neither the behavior manifold nor the operators
$P_T$ and $\Pi$ are available explicitly.
Section~\ref{sec:method} shows how the pretrained flow policy provides
approximations to both.

\section{RetraQ: Retraction-based Q-guidance}
\label{sec:method}

\begin{figure}[t]
    \centering
    \includegraphics[width=0.9\textwidth]{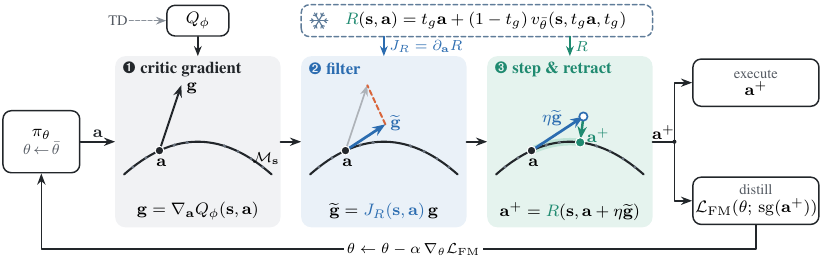}
    \caption{
    \textbf{One RetraQ update.}
    \stepA~At an action $\mathbf a$, the critic gradient $\mathbf g$ may
    contain components normal to the behavior manifold.
    \stepB~The Jacobian $J_R$ of the frozen flow's denoising map keeps tangent
    directions and suppresses normal ones.
    \stepC~After a step along the filtered direction $\widetilde{\mathbf g}$,
    the same map moves the action back toward the behavior manifold,
    giving $\mathbf a^+$. Repeating the update gives the refined action
    $\mathbf a^\star$, which is executed or distilled into the actor
    $\pi_\theta$ as a stop-gradient target.
    }
    \label{fig:retraq-method}
\end{figure}

Section~\ref{sec:identifiability} motivates two operations for critic guidance:
tangent filtering with $P_T$ and endpoint correction with $\Pi_{\mathbf s}$.
RetraQ approximates both using a frozen behavior-cloned (BC) flow.
In Eq.~\eqref{eq:refinement-template}, it sets
$F=J_R$ and $\mathcal C=R$, where $R$ is a denoising map of the flow and $J_R$
its Jacobian (Figure~\ref{fig:retraq-method}).

\subsection{The Flow's Denoising Map as Filter and Retraction}
\label{sec:retraq-refinement}

\textbf{The denoising map.}
\phantomsection\label{sec:flow-geometry}%
Let $\bar\theta$ denote the frozen parameters of the pretrained BC flow.
Dividing Eq.~\eqref{eq:fm-interpolation} by $t$ gives
$\mathbf x_t/t=\mathbf a+\sigma_t\mathbf z$ with $\sigma_t:=(1-t)/t$. Thus, after rescaling,
a flow state can be viewed as a clean action corrupted
by Gaussian noise. At a fixed
\emph{guidance level} $t_g\in(0,1)$, let $\sigma_g:=\sigma_{t_g}$. We interpret an input action $\mathbf a$ as the rescaled noisy observation
$\mathbf x_{t_g}/t_g$ and apply the clean-action
estimate of Eq.~\eqref{eq:clean-action-map}, which takes one forward pass:
\begin{equation}
    R(\mathbf s,\mathbf a):=D_{t_g}^{\bar\theta}(\mathbf s,t_g\mathbf a)
    =t_g\mathbf a+(1-t_g)\,v_{\bar\theta}(\mathbf s,t_g\mathbf a,t_g),
    \qquad
    J_R(\mathbf s,\mathbf a):=\frac{\partial R(\mathbf s,\mathbf a)}{\partial\mathbf a}\in\mathbb R^{D\times D}.
    \label{eq:retraq-reference-map}
\end{equation}

To see why $R$ and $J_R$ approximate these two operators, consider the
population-optimal velocity, for which the clean-action map is the posterior
mean of the clean action (Section~\ref{sec:prelim}).
Let $\mathbf A\sim\mu(\cdot\mid\mathbf s)$ and
$\mathbf Z\sim\mathcal N(0,I_D)$ be independent.
Then~\citep{manor2024posterior}
\begin{equation}
    R^{\mathrm{opt}}(\mathbf s,\mathbf a)=\mathbb E\bigl[\mathbf A\mid\mathbf A+\sigma_g\mathbf Z=\mathbf a\bigr],
    \qquad
    J_R^{\mathrm{opt}}(\mathbf s,\mathbf a)=\sigma_g^{-2}\operatorname{Cov}\bigl[\mathbf A\mid\mathbf A+\sigma_g\mathbf Z=\mathbf a\bigr].
    \label{eq:retraq-posterior-identity}
\end{equation}
Near the behavior manifold, this posterior concentrates around
$\Pi_{\mathbf s}(\mathbf a)$ and varies primarily along tangent directions,
suggesting $R\approx\Pi_{\mathbf s}$ and $J_R\approx P_T$.
Proposition~\ref{prop:retraq-local-geometry} (Appendix~\ref{app:retraq-theory})
makes this precise. For a compact $\mathcal M_{\mathbf s}$ with smooth positive density and for
points inside a tubular neighborhood of radius smaller than its reach,
\begin{equation}
    \underbrace{R^{\mathrm{opt}}(\mathbf s,\mathbf a)
      =\Pi_{\mathbf s}(\mathbf a)+O(\sigma_g^2)}_{\text{\stepC~retraction, }\mathcal C=\Pi_{\mathbf s}}
    \qquad\text{and, for }\mathbf a\in\mathcal M_{\mathbf s},\qquad
    \underbrace{J_R^{\mathrm{opt}}(\mathbf s,\mathbf a)
      =P_T(\mathbf a)+O(\sigma_g^2)}_{\text{\stepB~tangent filter, }F=P_T}.
    \label{eq:retraq-geometric-conditions}
\end{equation}
For a learned flow, these approximations incur additional error from the velocity field and its input derivatives
(Proposition~\ref{prop:retraq-learned-geometry}).
We use a fixed guidance level $t_g=0.9$ across all platforms.
Appendix~\ref{app:retraq-guidance-level} discusses this choice.

\subsection{The RetraQ Update}
\label{sec:retraq-update}

Replacing $P_T$ by $J_R$ and $\Pi_{\mathbf s}$ by $R$ in the ideal update of
Eq.~\eqref{eq:ideal-manifold-update} gives RetraQ. For a context $\mathbf s$, an
action $\mathbf a$, and a differentiable critic $Q$,
\begin{equation}
    \begin{alignedat}{3}
    &\stepA\quad&\mathbf g&=\nabla_{\mathbf a}Q(\mathbf s,\mathbf a)
      &\qquad&\text{critic ascent direction},\\
    &\stepB\quad&\widetilde{\mathbf g}&=\textcolor{retraqBlue}{J_R(\mathbf s,\mathbf a)}\,\mathbf g
      &&\text{filter the direction},\\
    &\stepC\quad&\mathbf a^+&=\textcolor{retraqGreen}{R}\bigl(\mathbf s,\mathbf a+\eta\,\widetilde{\mathbf g}\bigr)
      &&\text{step and retract}.
    \end{alignedat}
    \label{eq:retraq-step}
\end{equation}
The update needs no manifold estimate or
backpropagation through the sampling chain: step~\stepB{} is one Jacobian--vector product through $R$.
Applying Eq.~\eqref{eq:retraq-step} $M$ times to an initial 
action $\mathbf a^{(0)}$ gives the \emph{refined action}
$\mathbf a^\star=\textsc{Refine}(\mathbf s,\mathbf a^{(0)})$
(Algorithm~\ref{alg:retraq-training}); all experiments use $M=1$. The step
size $\eta$ is calibrated once
so that the critic-induced displacement of the first step matches a target
$d_{\mathrm{target}}$
(Appendix~\ref{app:retraq-update-details}).

\begin{algorithm}[t]
\caption{Offline-to-online RetraQ. \textsc{Refine} (lines~1--8) is shared by all
settings: test-time guidance runs only \textsc{Refine} and executes
$\mathbf a^\star$, and offline distillation skips
line~\ref{line:retraq-collect}. The full version is
Algorithm~\ref{alg:retraq-online-full}.}
\label{alg:retraq-training}
\small
\begin{algorithmic}[1]
\Require frozen BC flow $v_{\bar\theta}$; demonstrations $\mathcal D_{\mathrm{demo}}$; replay buffer $\mathcal D$;
         guidance level $t_g$; target displacement $d_{\mathrm{target}}$;
         refinement steps $M$; loss weights $\lambda_{\mathrm{imp}},\lambda_{\mathrm{BC}}$
\Function{Refine}{$\mathbf s,\mathbf a^{(0)};\,\hat Q,\eta$}
    \For{$\ell=0,\dots,M-1$}
        \State $\mathbf g\gets\nabla_{\mathbf a}\hat Q(\mathbf s,\mathbf a^{(\ell)})$
               \Comment{\stepA{} critic ascent direction}
        \State $\widetilde{\mathbf g}\gets J_R(\mathbf s,\mathbf a^{(\ell)})\,\mathbf g$
               \Comment{\stepB{} damp normal part; one JVP}
        \State $\mathbf a^{(\ell+1)}\gets\operatorname{sg}\bigl[R(\mathbf s,\mathbf a^{(\ell)}+\eta\,\widetilde{\mathbf g})\bigr]$
               \Comment{\stepC{} step, then retract}
    \EndFor
    \State \Return $\mathbf a^\star=\mathbf a^{(M)}$
\EndFunction
\Statex
\State $\theta\gets\bar\theta$;\; $\eta\gets$ unset
       \Comment{actor starts at BC; $\bar\theta$ stays frozen}
\State Warm up the critic on $\mathcal D_{\mathrm{demo}}$; actor fixed
       \Comment{no targets from an untrained critic}
\While{training budget remains}
    \State Add labeled rollouts of $\pi_\theta$ to $\mathcal D$
           \label{line:retraq-collect}\Comment{new data from current actor}
    \For{each update cycle}
        \State Update the critic on $\mathcal D$; freeze $\hat Q=\min_i Q_{\phi_i}$
               \Comment{fixed critic for this cycle}
        \For{each actor step of the cycle}
            \State Sample $\{(\mathbf s_j,\mathbf a^{\mathrm{exec}}_j)\}_j$ from $\mathcal D$
                   \Comment{executed chunks}
            \State \textbf{if} $\eta$ is unset \textbf{then} calibrate $\eta$ on this batch
                   \Comment{then held fixed}
            \State $\mathbf a^\star_j\gets\Call{Refine}{\mathbf s_j,\mathbf a^{\mathrm{exec}}_j;\,\hat Q,\eta}$;\;
                   $\mathbf a^{\mathrm{BC}}_j\sim\pi_{\bar\theta}(\cdot\mid\mathbf s_j)$
                   \Comment{target, anchor}
            \State Take a gradient step on $\mathcal L_{\mathrm{actor}}(\theta)$
                   \Comment{Eq.~\eqref{eq:retraq-distillation}}
        \EndFor
    \EndFor
\EndWhile
\end{algorithmic}
\end{algorithm}

\subsection{Critics and Policy Improvement}
\label{sec:retraq-deployment}
\phantomsection\label{sec:retraq-critic}\label{sec:retraq-data}%

RetraQ needs only the action gradient of a critic. The refined action
$\mathbf a^\star$ is a constant ($\operatorname{sg}[\cdot]$) that is either
executed or used as a regression target, in three settings that share
\textsc{Refine} (Table~\ref{tab:retraq-settings}).

\textbf{Test-time guidance.}
All parameters stay fixed, and $\mathbf a^\star$ is executed. Standard rollouts
record one action per state and so say little about how a slightly different
action would fare, which is what $\nabla_{\mathbf a}Q$ must capture. We
therefore collect \emph{forked rollouts}: at a few task-critical \emph{fork
states}, we replay the same prefix, perturb the chunk that the frozen policy
predicts there (the reference chunk $\mathbf a_c^{\mathrm{ref}}$ of fork group
$c$), and let the frozen policy finish the episode. Since branches share the
state, outcome differences can be attributed to the action. The fork critic
scores the deviation from the reference,
\begin{equation}
    Q_\phi(\mathbf s,\mathbf a)
    =\operatorname{sigmoid}\!\bigl(
        B_\psi(\mathbf s)+A_\omega(\mathbf a-\mathbf a_c^{\mathrm{ref}})
    \bigr),
    \qquad A_\omega(\mathbf 0)=0,
    \label{eq:fork-critic}
\end{equation}
so its action gradient comes entirely from $A_\omega$. Guidance is applied at
detected fork states; elsewhere the BC policy runs unchanged
(Appendix~\ref{app:retraq-data}).

\paragraph{Distillation.}
In the other two settings, an actor $\pi_\theta$, initialized at $\bar\theta$,
fits refined actions with the flow-matching loss
$\mathcal L_{\mathrm{FM}}(\theta;\mathbf s,\mathbf a)
:=\mathbb E_{t,\mathbf z}\|v_\theta(\mathbf s,\mathbf x_t,t)-(\mathbf a-\mathbf z)\|_2^2$
and is anchored to the frozen BC policy:
\begin{equation}
    \mathcal L_{\mathrm{actor}}(\theta)
    =\mathbb E_{\mathbf s}\bigl[
      \lambda_{\mathrm{imp}}\,\mathcal L_{\mathrm{FM}}(\theta;\mathbf s,\mathbf a^\star)
      +\lambda_{\mathrm{BC}}\,\mathcal L_{\mathrm{FM}}(\theta;\mathbf s,\mathbf a^{\mathrm{BC}})
    \bigr],
    \qquad
    \mathbf a^{\mathrm{BC}}\sim\pi_{\bar\theta}(\cdot\mid\mathbf s).
    \label{eq:retraq-distillation}
\end{equation}
\emph{Offline distillation} refines the current actor's actions on a fixed
dataset. \emph{Offline-to-online learning} (Algorithm~\ref{alg:retraq-training})
interleaves collecting rollouts into a replay buffer $\mathcal D$ with critic
and actor updates. Its twin TD critics (IQL~\citep{kostrikov2022iql} on the
robot) are regressed only at executed chunks, so their values are supported by
data while their normal gradients are not, exactly the situation of
Section~\ref{sec:identifiability}; refinement therefore starts from executed
chunks.


\section{Experiments}
\label{sec:experiments}

We evaluate RetraQ at three levels. Controlled problems with a known manifold
test the mechanism directly: whether the flow's Jacobian filters normal
directions, whether updates exploit the critic, and where the learned
geometry begins to fail (Section~\ref{sec:exp-geometry}). Simulation
benchmarks with component ablations test whether the two constraints improve
policies, and which one carries the gain (Section~\ref{sec:exp-simulation}).
A real-robot task tests the same update for test-time guidance, offline
distillation, and offline-to-online learning (Section~\ref{sec:exp-real}).

\subsection{Experimental Setup}
\label{sec:exp-setup}

We use controlled problems with a known manifold; three simulation
benchmarks, LIBERO~\citep{libero} (ten Goal and ten Object tasks, and the
six Spatial tasks with the lowest BC success, since the remaining Spatial
tasks leave little room for improvement), RoboMimic Can~\citep{robomimic}, and
OGBench cube-single~\citep{ogbench}; and a real-robot insertion task with a
$\pi_{0.5}$ policy~\citep{intelligence2025pi05}. Baselines are our
implementations of AWR~\citep{awr}, FlowDPG~\citep{flowdpg},
Q-VGM~\citep{qvgm_v1}, QAM~\citep{qam}, and, on the robot, QGF~\citep{qgf}. Component ablations
isolate the two constraints: \emph{raw ascent} $\mathbf a+\eta\mathbf g$,
\emph{filter-only} $\mathbf a+\eta J_R\mathbf g$, \emph{Raw$+R$}
$R(\mathbf a+\eta\mathbf g)$, and RetraQ $R(\mathbf a+\eta J_R\mathbf g)$; a
\emph{self-distillation} control refits the actor to its own actions. On
LIBERO, we report the final checkpoint, averaged over three actor seeds and
30 episodes per task and seed. Appendix~\ref{app:retraq-experiments} gives the
full protocols.

\subsection{Does the Flow Provide the Geometry, and Does It Prevent Exploitation?}
\label{sec:exp-geometry}

\begin{figure}[t]
\centering
\includegraphics[width=\linewidth]{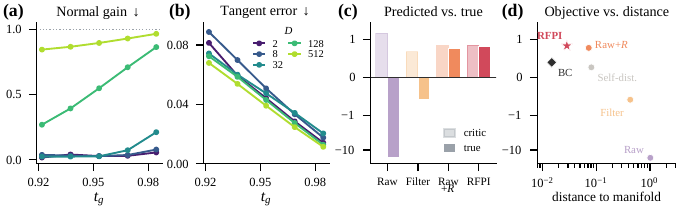}
\caption{\textbf{Controlled problems with a known manifold.}
\textbf{(a,\,b)}~Median normal gain ($0$: ideal filter; $1$: no filtering) and
tangent error of the learned $J_R$ versus the guidance level, for flat
manifolds of increasing dimension at a fixed flow-training budget.
\textbf{(c)}~$D=16$ problem: objective predicted by the critic (light) and
true objective (solid) after refinement. \textbf{(d)}~True objective versus the
95th-percentile distance to the manifold. Symmetric-log scales in (c,\,d);
details in Appendix~\ref{app:retraq-geometry-diagnostics}.}
\label{fig:retraq-geometry}
\end{figure}

A known manifold lets us measure what success rates cannot: whether $J_R$
filters normal directions, whether the critic is exploited, and whether refined
actions stay on the manifold (Figure~\ref{fig:retraq-geometry}).
\textbf{The flow's Jacobian acts as a tangent filter, with the trade-off
predicted by the theory.} For $D\le32$, the median normal gain stays below
$0.1$ at all but the highest guidance level, so $J_R$ removes over $90\%$ of
the normal component while keeping tangent directions (tangent error below
$0.09$). Raising $t_g$ lowers the tangent error but raises the normal gain
toward $1$, that is, $J_R\to I_D$, as
Proposition~\ref{prop:retraq-learned-geometry} predicts. The effect is strong at
$D=128$, and at $D=512$ filtering is largely lost under the fixed training
budget (normal gain $0.84$--$0.97$), where the flow also fits worst.
\textbf{The retraction closes the gap between predicted and true
improvement.} The critic predicts an improvement for every update. Raw ascent,
however, turns a predicted objective of $1.50$ into a true objective of
$-16.8$; filtering narrows this gap, but filter-only still turns $0.68$ into
$-0.58$. For both retracted updates, predicted and true objectives differ by at
most $0.07$ (Figure~\ref{fig:retraq-geometry}c).
\textbf{RetraQ attains the highest true objective and stays closest to the
manifold.} Its refined actions are $2.6\times$ closer to the manifold than
those of Raw$+R$ and $36\times$ closer than those of raw ascent
(Figure~\ref{fig:retraq-geometry}d), so with the retraction held fixed, the
filter further reduces the distance to the manifold.

\subsection{Policy Improvement in Simulation}
\label{sec:exp-simulation}

\begin{table}[t]
\centering
\small
\setlength{\tabcolsep}{6pt}
\caption{\textbf{Final online success (\%) on LIBERO}, mean $\pm$ standard
deviation over three actor seeds; actors train online from BC, and baselines
are our implementations. Best in bold.}
\label{tab:retraq-sim-main}
\begin{tabular}{@{}lccc@{}}
\toprule
Method & Spatial (6 tasks) & Goal (10 tasks) & Object (10 tasks)\\
\midrule
BC                & $63.3$ & $69.0$ & $55.3$\\
\midrule
AWR               & $64.8\pm0.6$ & $71.9\pm3.3$ & $58.4\pm0.7$\\
FlowDPG           & $60.4\pm3.6$ & $68.1\pm3.2$ & $49.3\pm8.0$\\
Q-VGM             & $70.6\pm0.6$ & $73.6\pm0.4$ & $59.9\pm2.0$\\
QAM               & $61.5\pm2.7$ & $73.1\pm0.7$ & $60.9\pm0.7$\\
\midrule
Self-distillation & $35.7\pm12.9$ & $59.6\pm11.9$ & $42.2\pm5.3$\\
Raw ascent        & $59.3\pm7.3$ & $67.9\pm5.8$ & $56.6\pm6.4$\\
Filter-only       & $63.5\pm10.1$ & $70.9\pm5.6$ & $57.8\pm6.0$\\
Raw$+R$           & $75.4\pm1.7$ & $79.4\pm1.4$ & $62.1\pm6.7$\\
\textbf{RetraQ}     & $\mathbf{78.1\pm4.2}$ & $\mathbf{81.3\pm3.7}$ & $\mathbf{66.4\pm5.2}$\\

\bottomrule
\end{tabular}
\end{table}

On LIBERO, actors start from BC and train online with a frozen geometry
reference (Table~\ref{tab:retraq-sim-main}).
\textbf{RetraQ improves BC on every suite and attains the highest mean
success}, $11$--$15$ points above BC and $5.5$--$7.8$ points above the strongest
baseline on each suite (Q-VGM on Spatial and Goal, QAM on Object), with three
seeds per arm. \textbf{The retraction carries most of the gain, and filtering adds to
it.} Without the retraction, filter-only trails RetraQ by $8.6$--$14.6$ points;
adding the retraction to raw ascent (Raw$+R$) recovers $5.6$--$16.1$ points,
and filtering adds another $1.9$--$4.3$ points on every suite, consistent in
sign but within the variation across three seeds. Self-distillation falls below
BC, so refitting the actor alone does not explain the improvement;
Appendix~\ref{app:retraq-behavior-reg} argues that behavior regularization does
not explain it either.

\textbf{What the filter changes differs across tasks.} On OGBench
cube-single, the filtered gradient stays almost parallel to the raw one
(median cosine $0.99$ throughout training) at $0.42$--$0.48$ of its norm.
Since the step is calibrated to a fixed displacement, the filter then mainly
rescales the step, and RetraQ and Raw$+R$ both approach the ceiling ($98.8\%$
and $99.0\%$). On RoboMimic Can, the filter also rotates the gradient (cosine
$0.67$--$0.71$), and RetraQ reaches the highest held-out success, $70.3\%$,
against $53.0\%$ for BC, $69.0\%$ for Q-VGM, and $67.0\%$ for Raw$+R$ (paired
$95\%$ interval for RetraQ minus Raw$+R$: $[-2.0,8.7]$ points).
An exploratory Can study that varies the critic's training set, with rounds
selected on the evaluation episodes, suggests that RetraQ benefits more from a
stronger critic (Appendix~\ref{app:retraq-critic-quality}).

\subsection{Real-Robot Guidance and Online Improvement}
\label{sec:exp-real}

\begin{figure}[t]
\centering
\includegraphics[width=\linewidth]{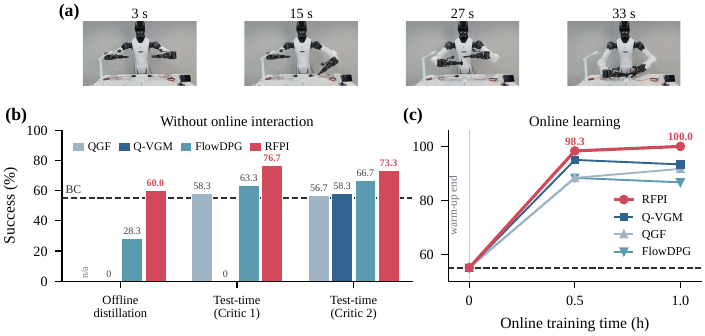}
\caption{\textbf{Real-robot plug insertion.}
\textbf{(a)}~A successful rollout of the online RetraQ policy.
\textbf{(b)}~Without online interaction: offline distillation, and test-time
guidance of the frozen BC policy with two fork critics (n/a: not applicable). \textbf{(c)}~Offline-to-online learning from a shared BC policy and
critic warm-up; means over three training runs. Numbers in Appendix~\ref{app:retraq-real-robot}.}
\label{fig:retraq-real}
\end{figure}

We evaluate phone-charging plug insertion on an AgiBot G02 dual-arm robot with
a behavior-cloned $\pi_{0.5}$ policy that succeeds in $55\%$ of trials; every
evaluation uses 60 trials (Appendix~\ref{app:retraq-real-robot}).

\textbf{Test-time guidance.} Without any parameter update, RetraQ raises success
to $76.7\%$ and $73.3\%$ with two fork critics, linear and nonlinear in the
action deviation, each trained on 270 forked rollouts. All guidance methods
take steps of the same median size, so they differ only in direction and
destination; the best baseline, FlowDPG, reaches $63.3\%$ and $66.7\%$, and
Q-VGM falls to $0\%$ with the linear critic (Figure~\ref{fig:retraq-real}b).
With 60 trials per arm, RetraQ's gain over BC is significant with Critic~1
(Fisher's exact test, $p=0.02$; $p=0.06$ with Critic~2), whereas its margin
over FlowDPG is not ($p\ge0.16$). Appendix~\ref{app:retraq-real-robot} analyzes
Q-VGM's failure.

\textbf{Offline distillation.} Distilling a fixed set of refined targets under
one shared, large displacement budget separates the methods further. Q-VGM
targets drive success to $0\%$, with the policy repeatedly hitting the table,
and FlowDPG targets to $28.3\%$, with frequent misaligned insertions, whereas
the RetraQ-distilled policy reaches $60.0\%$, on par with BC ($p=0.71$).

\textbf{Offline-to-online learning.} All methods start from the same BC policy
and warmed-up IQL critic, and then train their own critics under the same
interaction budget; results are means over three training runs per method,
each checkpoint evaluated with 60 trials per run.
RetraQ reaches the highest success at every checkpoint
(Figure~\ref{fig:retraq-real}c): $98.3\%$ after 30 minutes and $100\%$ after one
hour, against at most $95.0\%$ and $93.3\%$ for the baselines.

\section{Related Work}
\label{sec:related}

\paragraph{Critic-guided improvement of generative policies.}
A learned critic can improve a diffusion or flow policy through its values
alone~\citep{awr,kostrikov2022iql,zhang2025qipo}, by backpropagating
$\nabla_{\mathbf a}Q$ through the generation
process~\citep{wang2023diffusionql,park2025fql,qam}, or by applying
$\nabla_{\mathbf a}Q$ locally at one action without differentiating through
sampling~\citep{yang2023dipo,flowdpg,qvgm_v1,qgf}. The local methods, closest to
RetraQ, follow the gradient unfiltered and correct the endpoint at most by
clipping or best-iterate selection, that is, $F=I_D$ and $\mathcal C\approx\mathrm{id}$
in Eq.~\eqref{eq:refinement-template}. RetraQ instead uses the frozen flow's
denoising map as both operators, $F=J_R$ and $\mathcal C=R$, and so acts on
exactly the normal gradient component that the data leave undetermined
(Section~\ref{sec:identifiability}). Offline RL counters the same extrapolation
by constraining the policy or its values~\citep{fujimoto2019offpolicy,kumar2019bear};
RetraQ constrains each update geometrically, with the behavior model already at
hand.

\paragraph{Riemannian optimization through denoising.}
The ideal update of Eq.~\eqref{eq:ideal-manifold-update} is Riemannian gradient
ascent with the projection retraction~\citep{absil2012projection,boumal}, and a
denoiser's Jacobian is a scaled posterior
covariance~\citep{efron,manor2024posterior}. Landing with the
Score~\citep{landingscore} combines the two into denoising Riemannian gradient
descent (DRGD) for a known objective and a pretrained diffusion denoiser.
RetraQ shares this filter--step--denoise structure but differs in each
ingredient: it follows a \emph{learned} critic, for which filtering is
motivated by the non-identifiability of normal gradients
(Section~\ref{sec:identifiability}); it obtains $R$ and $J_R$ from one velocity
evaluation of a context-conditioned flow policy by rescaling flow states into
Gaussian-corrupted actions (Section~\ref{sec:retraq-refinement}); it bounds how
errors of a learned velocity and its input derivatives enter both operators
(Proposition~\ref{prop:retraq-learned-geometry}); and it uses refined actions as
distillation and online-learning targets. Manifold-preserving diffusion
guidance~\citep{mpgd} and latent-noise steering~\citep{dsrl} are compared in
Appendix~\ref{app:related-work}.
\section{Conclusion}
\label{sec:conclusion}

Accurate critic values on behavior data leave gradients normal to the
behavior manifold undetermined. RetraQ uses the pretrained flow to constrain
both the direction and the destination of critic-guided updates: the
Jacobian of its denoising map filters the critic gradient, and the map
itself brings the updated action back toward the manifold, approximating
tangent projection and retraction. In our experiments, RetraQ
outperforms the evaluated baselines across tested LIBERO suites.
On real plug insertion, it improves a $\pi_{0.5}$ policy at test time
and reaches near-perfect success within an hour of online training.
These results show that a pretrained generative
policy provides not only an expressive action distribution, but also local
geometry for reliable policy improvement.

\paragraph{Limitations and future work.}
\phantomsection\label{sec:limitations}%
Repeated offline distillation without new interaction is a limitation: on
LIBERO, RetraQ degrades over three refinement rounds more than AWR or FlowDPG
(Appendix~\ref{sec:exp-boundaries}), so its stability advantage holds against
raw ascent but not in general. Separating errors in the refined targets from
errors in fitting them could help stabilize offline improvement. The Jacobian--vector product through the frozen flow also adds
GPU memory.
Our geometric guarantees are local and asymptotic, and the learned filter
weakens in high-dimensional action spaces; constraining velocity derivatives
during training or adapting the guidance level may help. Refinement also remains tied to the
support of the frozen reference flow; combining RetraQ with exploration
or carefully updating the reference could broaden the behaviors it
can learn. 

\section*{AI Use Statement}
 
In this work, we used generative AI tools (large language model assistants)
to assist in creating the plotting code for the figures, formatting and checking
references. We have not used generative AI tools to generate data sets,
to collect or label real-robot data, or to translate text.
We take responsibility for the final content of this work.
 
\section*{Reproducibility Statement}
 
Appendix~\ref{app:geometric-refinement} states the assumptions of every
theoretical result and gives complete proofs.
Appendix~\ref{app:retraq-implementation} specifies the operator, step-size
calibration, critics, and all three settings, including the full
offline-to-online algorithm (Algorithm~\ref{alg:retraq-online-full}) and the
hyperparameters on each platform (Table~\ref{tab:retraq-hparams}).
Appendix~\ref{app:retraq-experiments} gives, for each benchmark, the task
subsets, seeds, evaluation episodes, checkpoint selection, and all recorded
arms, as well as the real-robot platform, policy, and trial protocol.
Anonymized code for the algorithms will be released.

\bibliography{iclr2027_conference}

\begin{thebibliography}{35}
\providecommand{\natexlab}[1]{#1}
\providecommand{\url}[1]{\texttt{#1}}
\expandafter\ifx\csname urlstyle\endcsname\relax
  \providecommand{\doi}[1]{doi: #1}\else
  \providecommand{\doi}{doi: \begingroup \urlstyle{rm}\Url}\fi

\bibitem[Absil \& Malick(2012)Absil and Malick]{absil2012projection}
P.-A. Absil and J{\'e}r{\^o}me Malick.
\newblock Projection-like retractions on matrix manifolds.
\newblock \emph{SIAM Journal on Optimization}, 22\penalty0 (1):\penalty0
  135--158, 2012.
\newblock \doi{10.1137/100802529}.

\bibitem[Black et~al.(2025{\natexlab{a}})Black, Brown, Darpinian, Dhabalia,
  Driess, Esmail, Equi, Finn, Fusai, Galliker, Ghosh, Groom, Hausman, Ichter,
  Jakubczak, Jones, Ke, LeBlanc, Levine, Li-Bell, Mothukuri, Nair, Pertsch,
  Ren, Shi, Smith, Springenberg, Stachowicz, Tanner, Vuong, Walke, Walling,
  Wang, Yu, and Zhilinsky]{intelligence2025pi05}
Kevin Black, Noah Brown, James Darpinian, Karan Dhabalia, Danny Driess, Adnan
  Esmail, Michael~Robert Equi, Chelsea Finn, Niccolo Fusai, Manuel~Y. Galliker,
  Dibya Ghosh, Lachy Groom, Karol Hausman, Brian Ichter, Szymon Jakubczak, Tim
  Jones, Liyiming Ke, Devin LeBlanc, Sergey Levine, Adrian Li-Bell, Mohith
  Mothukuri, Suraj Nair, Karl Pertsch, Allen~Z. Ren, Lucy~Xiaoyang Shi, Laura
  Smith, Jost~Tobias Springenberg, Kyle Stachowicz, James Tanner, Quan Vuong,
  Homer Walke, Anna Walling, Haohuan Wang, Lili Yu, and Ury Zhilinsky.
\newblock {$\pi_{0.5}$}: A vision-language-action model with open-world
  generalization.
\newblock In \emph{Proceedings of the 9th Conference on Robot Learning}, volume
  305 of \emph{Proceedings of Machine Learning Research}, pp.\  17--40. PMLR,
  2025{\natexlab{a}}.

\bibitem[Black et~al.(2025{\natexlab{b}})Black, Brown, Driess, Esmail, Equi,
  Finn, Fusai, Groom, Hausman, Ichter, Jakubczak, Jones, Ke, Levine, Li-Bell,
  Mothukuri, Nair, Pertsch, Shi, Smith, Tanner, Vuong, Walling, Wang, and
  Zhilinsky]{black2024pi0}
Kevin Black, Noah Brown, Danny Driess, Adnan Esmail, Michael~Robert Equi,
  Chelsea Finn, Niccolo Fusai, Lachy Groom, Karol Hausman, Brian Ichter, Szymon
  Jakubczak, Tim Jones, Liyiming Ke, Sergey Levine, Adrian Li-Bell, Mohith
  Mothukuri, Suraj Nair, Karl Pertsch, Lucy~Xiaoyang Shi, Laura Smith, James
  Tanner, Quan Vuong, Anna Walling, Haohuan Wang, and Ury Zhilinsky.
\newblock {$\pi_0$}: A vision-language-action flow model for general robot
  control.
\newblock In \emph{Proceedings of Robotics: Science and Systems},
  2025{\natexlab{b}}.
\newblock \doi{10.15607/RSS.2025.XXI.010}.

\bibitem[Boumal(2023)]{boumal}
Nicolas Boumal.
\newblock \emph{An Introduction to Optimization on Smooth Manifolds}.
\newblock Cambridge University Press, 2023.
\newblock \doi{10.1017/9781009166164}.

\bibitem[D'Oro \& Ja{\'s}kowski(2020)D'Oro and Ja{\'s}kowski]{mage}
Pierluca D'Oro and Wojciech Ja{\'s}kowski.
\newblock How to learn a useful critic? {M}odel-based action-gradient-estimator
  policy optimization.
\newblock In \emph{Advances in Neural Information Processing Systems},
  volume~33, pp.\  313--324. Curran Associates, Inc., 2020.

\bibitem[Efron(2011)]{efron}
Bradley Efron.
\newblock {Tweedie}'s formula and selection bias.
\newblock \emph{Journal of the American Statistical Association}, 106\penalty0
  (496):\penalty0 1602--1614, 2011.
\newblock \doi{10.1198/jasa.2011.tm11181}.

\bibitem[Federer(1959)]{federer}
Herbert Federer.
\newblock Curvature measures.
\newblock \emph{Transactions of the American Mathematical Society}, 93\penalty0
  (3):\penalty0 418--491, 1959.
\newblock \doi{10.1090/S0002-9947-1959-0110078-1}.
\newblock URL
  \url{https://www.ams.org/tran/1959-093-03/S0002-9947-1959-0110078-1/}.

\bibitem[Florence et~al.(2022)Florence, Lynch, Zeng, Ramirez, Wahid, Downs,
  Wong, Lee, Mordatch, and Tompson]{florence2022ibc}
Pete Florence, Corey Lynch, Andy Zeng, Oscar~A. Ramirez, Ayzaan Wahid, Laura
  Downs, Adrian Wong, Johnny Lee, Igor Mordatch, and Jonathan Tompson.
\newblock Implicit behavioral cloning.
\newblock In \emph{Proceedings of the 5th Conference on Robot Learning}, volume
  164 of \emph{Proceedings of Machine Learning Research}, pp.\  158--168. PMLR,
  2022.

\bibitem[Fujimoto et~al.(2019)Fujimoto, Meger, and
  Precup]{fujimoto2019offpolicy}
Scott Fujimoto, David Meger, and Doina Precup.
\newblock Off-policy deep reinforcement learning without exploration.
\newblock In \emph{Proceedings of the 36th International Conference on Machine
  Learning}, volume~97 of \emph{Proceedings of Machine Learning Research}, pp.\
   2052--2062. PMLR, 2019.

\bibitem[He et~al.(2024)He, Murata, Lai, Takida, Uesaka, Kim, Liao, Mitsufuji,
  Kolter, Salakhutdinov, and Ermon]{mpgd}
Yutong He, Naoki Murata, Chieh-Hsin Lai, Yuhta Takida, Toshimitsu Uesaka,
  Dongjun Kim, Wei-Hsiang Liao, Yuki Mitsufuji, J.~Zico Kolter, Ruslan
  Salakhutdinov, and Stefano Ermon.
\newblock Manifold preserving guided diffusion.
\newblock In \emph{International Conference on Learning Representations}, 2024.

\bibitem[Kharitenko et~al.(2026)Kharitenko, Shen, {De Santi}, He, and
  D{\"o}rfler]{landingscore}
Andrey Kharitenko, Zebang Shen, Riccardo {De Santi}, Niao He, and Florian
  D{\"o}rfler.
\newblock Landing with the score: {R}iemannian optimization through denoising.
\newblock In \emph{International Conference on Learning Representations}, 2026.
\newblock URL \url{https://openreview.net/forum?id=xZNoeX0z9f}.

\bibitem[Kostrikov et~al.(2022)Kostrikov, Nair, and Levine]{kostrikov2022iql}
Ilya Kostrikov, Ashvin Nair, and Sergey Levine.
\newblock Offline reinforcement learning with implicit {Q}-learning.
\newblock In \emph{International Conference on Learning Representations}, 2022.
\newblock URL \url{https://openreview.net/forum?id=68n2s9ZJWF8}.

\bibitem[Kumar et~al.(2019)Kumar, Fu, Soh, Tucker, and Levine]{kumar2019bear}
Aviral Kumar, Justin Fu, Matthew Soh, George Tucker, and Sergey Levine.
\newblock Stabilizing off-policy {Q}-learning via bootstrapping error
  reduction.
\newblock In \emph{Advances in Neural Information Processing Systems},
  volume~32, pp.\  11761--11771. Curran Associates, Inc., 2019.

\bibitem[Li \& Levine(2026)Li and Levine]{qam}
Qiyang Li and Sergey Levine.
\newblock {Q}-learning with adjoint matching.
\newblock In \emph{International Conference on Learning Representations}, 2026.
\newblock URL \url{https://openreview.net/forum?id=vd4eNAdtO6}.

\bibitem[Lipman et~al.(2023)Lipman, Chen, Ben-Hamu, Nickel, and
  Le]{lipman2023flow}
Yaron Lipman, Ricky T.~Q. Chen, Heli Ben-Hamu, Maximilian Nickel, and Matt Le.
\newblock Flow matching for generative modeling.
\newblock In \emph{International Conference on Learning Representations}, 2023.
\newblock URL \url{https://openreview.net/forum?id=PqvMRDCJT9t}.

\bibitem[Liu et~al.(2023{\natexlab{a}})Liu, Zhu, Gao, Feng, Liu, Zhu, and
  Stone]{libero}
Bo~Liu, Yifeng Zhu, Chongkai Gao, Yihao Feng, Qiang Liu, Yuke Zhu, and Peter
  Stone.
\newblock {LIBERO}: Benchmarking knowledge transfer for lifelong robot
  learning.
\newblock In \emph{Advances in Neural Information Processing Systems},
  volume~36, pp.\  44776--44791. Curran Associates, Inc., 2023{\natexlab{a}}.
\newblock \doi{10.52202/075280-1939}.

\bibitem[Liu et~al.(2023{\natexlab{b}})Liu, Gong, and Liu]{liu2023rectified}
Xingchao Liu, Chengyue Gong, and Qiang Liu.
\newblock Flow straight and fast: Learning to generate and transfer data with
  rectified flow.
\newblock In \emph{International Conference on Learning Representations},
  2023{\natexlab{b}}.
\newblock URL \url{https://openreview.net/forum?id=XVjTT1nw5z}.

\bibitem[Luo et~al.(2025)Luo, Xu, Wu, and Levine]{luo2024hilserl}
Jianlan Luo, Charles Xu, Jeffrey Wu, and Sergey Levine.
\newblock Precise and dexterous robotic manipulation via human-in-the-loop
  reinforcement learning.
\newblock \emph{Science Robotics}, 10\penalty0 (105):\penalty0 eads5033, 2025.
\newblock \doi{10.1126/scirobotics.ads5033}.

\bibitem[Mandlekar et~al.(2022)Mandlekar, Xu, Wong, Nasiriany, Wang, Kulkarni,
  Fei-Fei, Savarese, Zhu, and Mart{\'i}n-Mart{\'i}n]{robomimic}
Ajay Mandlekar, Danfei Xu, Josiah Wong, Soroush Nasiriany, Chen Wang, Rohun
  Kulkarni, Li~Fei-Fei, Silvio Savarese, Yuke Zhu, and Roberto
  Mart{\'i}n-Mart{\'i}n.
\newblock What matters in learning from offline human demonstrations for robot
  manipulation.
\newblock In \emph{Proceedings of the 5th Conference on Robot Learning}, volume
  164 of \emph{Proceedings of Machine Learning Research}, pp.\  1678--1690.
  PMLR, 2022.

\bibitem[Manor \& Michaeli(2024)Manor and Michaeli]{manor2024posterior}
Hila Manor and Tomer Michaeli.
\newblock On the posterior distribution in denoising: Application to
  uncertainty quantification.
\newblock In \emph{International Conference on Learning Representations}, 2024.
\newblock URL \url{https://openreview.net/forum?id=adSGeugiuj}.

\bibitem[{NVIDIA} et~al.(2025){NVIDIA}, Bjorck, Casta{\~n}eda, Cherniadev, Da,
  Ding, Fan, Fang, Fox, Hu, Huang, et~al.]{bjorck2025gr00t}
{NVIDIA}, Johan Bjorck, Fernando Casta{\~n}eda, Nikita Cherniadev, Xingye Da,
  Runyu Ding, Linxi Fan, Yu~Fang, Dieter Fox, Fengyuan Hu, Spencer Huang,
  et~al.
\newblock {GR00T N1}: An open foundation model for generalist humanoid robots.
\newblock \emph{arXiv preprint arXiv:2503.14734}, 2025.

\bibitem[Park et~al.(2025{\natexlab{a}})Park, Frans, Eysenbach, and
  Levine]{ogbench}
Seohong Park, Kevin Frans, Benjamin Eysenbach, and Sergey Levine.
\newblock {OGBench}: Benchmarking offline goal-conditioned {RL}.
\newblock In \emph{International Conference on Learning Representations},
  2025{\natexlab{a}}.
\newblock URL \url{https://openreview.net/forum?id=M992mjgKzI}.

\bibitem[Park et~al.(2025{\natexlab{b}})Park, Li, and Levine]{park2025fql}
Seohong Park, Qiyang Li, and Sergey Levine.
\newblock Flow {Q}-learning.
\newblock In \emph{Proceedings of the 42nd International Conference on Machine
  Learning}, volume 267 of \emph{Proceedings of Machine Learning Research},
  pp.\  48104--48127. PMLR, 2025{\natexlab{b}}.

\bibitem[Peng et~al.(2019)Peng, Kumar, Zhang, and Levine]{awr}
Xue~Bin Peng, Aviral Kumar, Grace Zhang, and Sergey Levine.
\newblock Advantage-weighted regression: Simple and scalable off-policy
  reinforcement learning.
\newblock \emph{arXiv preprint arXiv:1910.00177}, 2019.

\bibitem[{Physical Intelligence} et~al.(2025){Physical Intelligence}, Amin,
  Aniceto, Balakrishna, Black, Conley, Connors, Darpinian, Dhabalia, DiCarlo,
  et~al.]{intelligence2025pistar06}
{Physical Intelligence}, Ali Amin, Raichelle Aniceto, Ashwin Balakrishna, Kevin
  Black, Ken Conley, Grace Connors, James Darpinian, Karan Dhabalia, Jared
  DiCarlo, et~al.
\newblock {$\pi^{*}_{0.6}$}: A {VLA} that learns from experience.
\newblock \emph{arXiv preprint arXiv:2511.14759}, 2025.

\bibitem[Psenka et~al.(2024)Psenka, Escontrela, Abbeel, and Ma]{psenka2024qsm}
Michael Psenka, Alejandro Escontrela, Pieter Abbeel, and Yi~Ma.
\newblock Learning a diffusion model policy from rewards via {Q}-score
  matching.
\newblock In \emph{Proceedings of the 41st International Conference on Machine
  Learning}, volume 235 of \emph{Proceedings of Machine Learning Research},
  pp.\  41163--41182. PMLR, 2024.

\bibitem[Shafiullah et~al.(2022)Shafiullah, Cui, Altanzaya, and
  Pinto]{shafiullah2022bet}
Nur Muhammad~Mahi Shafiullah, Zichen~Jeff Cui, Ariuntuya Altanzaya, and Lerrel
  Pinto.
\newblock Behavior transformers: Cloning $k$ modes with one stone.
\newblock In \emph{Advances in Neural Information Processing Systems},
  volume~35, pp.\  22955--22968. Curran Associates, Inc., 2022.
\newblock \doi{10.52202/068431-1668}.

\bibitem[Shi et~al.(2026)Shi, Shi, Hebbar, Zhao, Amarnath, Su, Bahl, and
  Pathak]{flowdpg}
Kexin Shi, Junyao Shi, Poorvi Hebbar, Zhuolun Zhao, Tarun Amarnath, Yifan Su,
  Shikhar Bahl, and Deepak Pathak.
\newblock {FlowDPG}: Deterministic policy gradient on flow matching policies
  for real-world manipulation.
\newblock \emph{arXiv preprint arXiv:2606.22303}, 2026.

\bibitem[Silver et~al.(2014)Silver, Lever, Heess, Degris, Wierstra, and
  Riedmiller]{silver2014dpg}
David Silver, Guy Lever, Nicolas Heess, Thomas Degris, Daan Wierstra, and
  Martin Riedmiller.
\newblock Deterministic policy gradient algorithms.
\newblock In \emph{Proceedings of the 31st International Conference on Machine
  Learning}, volume~32 of \emph{Proceedings of Machine Learning Research}, pp.\
   387--395. PMLR, 2014.

\bibitem[Wagenmaker et~al.(2025)Wagenmaker, Nakamoto, Zhang, Park, Yagoub,
  Nagabandi, Gupta, and Levine]{dsrl}
Andrew Wagenmaker, Mitsuhiko Nakamoto, Yunchu Zhang, Seohong Park, Waleed
  Yagoub, Anusha Nagabandi, Abhishek Gupta, and Sergey Levine.
\newblock Steering your diffusion policy with latent space reinforcement
  learning.
\newblock In \emph{Proceedings of the 9th Conference on Robot Learning}, volume
  305 of \emph{Proceedings of Machine Learning Research}, pp.\  258--282, 2025.

\bibitem[Wang et~al.(2023)Wang, Hunt, and Zhou]{wang2023diffusionql}
Zhendong Wang, Jonathan~J. Hunt, and Mingyuan Zhou.
\newblock Diffusion policies as an expressive policy class for offline
  reinforcement learning.
\newblock In \emph{International Conference on Learning Representations}, 2023.
\newblock URL \url{https://openreview.net/forum?id=AHvFDPi-FA}.

\bibitem[Wang et~al.(2026)Wang, Sun, Mao, Wang, and Mu]{qvgm_v1}
Ziqian Wang, Jiayu Sun, Xingjian Mao, Minqian Wang, and Yao Mu.
\newblock {Q-VGM}: {Q}-guided value-gradient matching for flow-matching {VLA}
  policies.
\newblock \emph{arXiv preprint arXiv:2606.08015v1}, 2026.
\newblock URL \url{https://arxiv.org/abs/2606.08015v1}.

\bibitem[Yang et~al.(2023)Yang, Huang, Lei, Zhong, Yang, Fang, Wen, Zhou, and
  Lin]{yang2023dipo}
Long Yang, Zhixiong Huang, Fenghao Lei, Yucun Zhong, Yiming Yang, Cong Fang,
  Shiting Wen, Binbin Zhou, and Zhouchen Lin.
\newblock Policy representation via diffusion probability model for
  reinforcement learning.
\newblock \emph{arXiv preprint arXiv:2305.13122}, 2023.

\bibitem[Zhang et~al.(2025)Zhang, Zhang, and Gu]{zhang2025qipo}
Shiyuan Zhang, Weitong Zhang, and Quanquan Gu.
\newblock Energy-weighted flow matching for offline reinforcement learning.
\newblock In \emph{International Conference on Learning Representations}, 2025.
\newblock URL \url{https://openreview.net/forum?id=HA0oLUvuGI}.

\bibitem[Zhou et~al.(2026)Zhou, Peng, Xu, Li, Springenberg, Frans, and
  Levine]{qgf}
Zhiyuan Zhou, Andy Peng, Charles Xu, Qiyang Li, Tobias Springenberg, Kevin
  Frans, and Sergey Levine.
\newblock Test-time gradient guidance of flow policies in reinforcement
  learning.
\newblock \emph{arXiv preprint arXiv:2606.11087}, 2026.

\end{thebibliography}
\bibliographystyle{iclr2027_conference}

\appendix

\section{Extended Related Work}
\label{app:related-work}

This appendix extends Section~\ref{sec:related}. We relate RetraQ to prior work through the template of
Eq.~\eqref{eq:refinement-template},
$\mathbf a^+=\mathcal C\bigl(\mathbf a+\eta F(\mathbf a)\mathbf g(\mathbf a)\bigr)$
with $\mathbf g=\nabla_{\mathbf a}Q$: Appendix~\ref{app:rw-critic} covers
critic-guided improvement of diffusion and flow policies, and
Appendix~\ref{app:rw-geometry} covers the Riemannian-optimization view of
the direction transform $F$ and the endpoint correction $\mathcal C$.

\subsection{Critic-Guided Improvement of Flow Policies}
\label{app:rw-critic}

\paragraph{Other uses of the critic.}
Value-based methods use the critic only through its values, for example by
advantage-weighted regression~\citep{awr,kostrikov2022iql} or by weighting the
flow-matching loss with $\exp(\beta Q)$~\citep{zhang2025qipo}; they never
differentiate $Q$ with respect to the action. Pathwise methods backpropagate
$\nabla_{\mathbf a}Q$ through the policy's generation process, either through
the full sampling chain~\citep{wang2023diffusionql}, a distilled one-step
actor~\citep{park2025fql}, or an adjoint of the behavior
field~\citep{qam}. Our experiments include AWR from the first group and QAM from the second.
Latent-steering methods such as DSRL~\citep{dsrl} leave the policy weights
fixed and instead run RL over the input noise of the generative policy; they
change which samples the policy produces rather than refining a given action
with the critic gradient, and are complementary to RetraQ.

\paragraph{Local critic-gradient guidance.}
The methods closest to RetraQ apply $\mathbf g=\nabla_{\mathbf a}Q$ at a single
action without differentiating through the sampling process
(Table~\ref{tab:rw-template}). DIPO fits the policy to raw-ascent targets
$\mathbf a+\eta\mathbf g$~\citep{yang2023dipo}. FlowDPG and Q-VGM evaluate
$\mathbf g$ at the one-step clean-action estimate $D_t(\mathbf s,\mathbf x_t)$
of Eq.~\eqref{eq:clean-action-map} and regress the velocity toward the
ascended action, with a step scaled to the velocity norm~\citep{flowdpg} or
with several ascent steps that are clipped to the action range and keep the
highest-valued iterate~\citep{qvgm_v1}. QGF adds the
gradient at $D_t$ to the velocity during test-time sampling, replacing the
transposed Jacobian required by the chain rule with the identity, which it
reports to outperform the full Jacobian~\citep{qgf}. In all of these, $F$ is the
identity up to a scalar or clipping, and $\mathcal C$ is at most coordinate
clipping or a value-based acceptance rule; $D_t$ only locates where the critic
is evaluated. RetraQ instead uses the frozen flow's map as the operators
themselves, $F=J_R$ and $\mathcal C=R$, which approximate tangent filtering
and retraction onto the behavior manifold (Section~\ref{sec:flow-geometry}).
Section~\ref{sec:experiments} compares our implementations of FlowDPG, Q-VGM,
and, on the robot, QGF; because the methods also differ in regularization and
data collection, these comparisons evaluate complete procedures rather than
the operators alone.

\begin{table}[h]
\centering
\small
\caption{Local critic-gradient methods in the template of
Eq.~\eqref{eq:refinement-template},
$\mathbf a^+=\mathcal C(\mathbf a+\eta F\mathbf g)$, as described in the
cited papers. $D_t(\mathbf s,\mathbf x_t)$ is the one-step clean-action
estimate of Eq.~\eqref{eq:clean-action-map} at the flow state $\mathbf x_t$;
$\mathrm{id}$ denotes no endpoint correction. The last three rows are the
operator variants of Section~\ref{sec:exp-setup}.
$^\dagger$QGF adds its guidance term to the velocity at $\mathbf x_t$ during
sampling.}
\label{tab:rw-template}
\begin{tabular}{@{}llll@{}}
\toprule
Method & $\mathbf g$ evaluated at & Direction $F$ & Endpoint $\mathcal C$\\
\midrule
DIPO & replay action & $I_D$ & $\mathrm{id}$\\
FlowDPG & $D_t(\mathbf s,\mathbf x_t)$ & $I_D$, step scaled to velocity norm & $\mathrm{id}$\\
Q-VGM & $D_t(\mathbf s,\mathbf x_t)$ & $I_D$, gradient clipped & action-range clip, best-$Q$ iterate\\
QGF$^\dagger$ & $D_t(\mathbf s,\mathbf x_t)$ & $I_D$ in place of $J^\top$ & $\mathrm{id}$\\
\midrule
Raw ascent & clean action $\mathbf a$ & $I_D$ & $\mathrm{id}$\\
Raw+$R$ & clean action $\mathbf a$ & $I_D$ & $R$\\
RetraQ & clean action $\mathbf a$ & $J_R(\mathbf a)$ & $R$\\
\bottomrule
\end{tabular}
\end{table}

\subsection{Riemannian Optimization through Denoising}
\label{app:rw-geometry}

\paragraph{Retraction and denoising.}
Eq.~\eqref{eq:ideal-manifold-update} is one step of Riemannian gradient
ascent with the metric-projection retraction~\citep{absil2012projection,boumal}.
On the manifold, the Jacobian of $\Pi$ is $P_T$~\citep[Sec.~5.12]{boumal}, so
a single map $R\approx\Pi$ can supply both $\mathcal C=R$ and $F=J_R$. The
flow's denoising map is such a map. By Tweedie's formula and its
second-order form~\citep{efron,manor2024posterior}, it is a posterior mean
whose Jacobian is the scaled posterior covariance
(Eq.~\eqref{eq:retraq-posterior-identity}); near a data manifold, these approach
$\Pi$ and its Jacobian~\citep{landingscore}, as stated in
Eq.~\eqref{eq:retraq-geometric-conditions}.

\paragraph{Manifold-preserving guidance.}
Training-free guidance of diffusion models also uses the manifold hypothesis:
MPGD~\citep{mpgd} applies the guidance gradient to the clean-sample estimate
and keeps the guided sample on the data manifold, for example with a
pretrained autoencoder. These methods steer the \emph{generation} of a sample
toward a known guidance objective. RetraQ instead refines a completed action with
a learned critic, motivates filtering by the non-identifiability of the critic's
normal gradients, obtains both operators from the frozen policy itself without
an auxiliary model, and uses the refined actions as policy-improvement targets.

\paragraph{Denoising Riemannian gradient descent.}
The closest predecessor of RetraQ's update is \emph{Landing with the
Score}~\citep{landingscore}, whose Denoising Riemannian Gradient
Descent~(DRGD) minimizes a known objective $f$ over a data manifold by
$\mathbf x_{k+1}=\mathbf v\bigl(\mathbf x_k-\gamma_k\,\mathbf v'(\mathbf x_k)\nabla f(\mathbf x_k)\bigr)$,
with $\mathbf v$ a pretrained diffusion denoiser. RetraQ's update in
Eq.~\eqref{eq:retraq-step} shares this filter--step--denoise structure, and we
view DRGD as its closest algorithmic predecessor. The two methods, however,
address different problems, and RetraQ changes how each ingredient is obtained
and justified.
\emph{First}, DRGD minimizes a known objective $f$, whose full gradient is
meaningful everywhere, and uses the denoiser for feasibility. RetraQ follows a
\emph{learned} critic whose values are constrained only on behavior data;
Section~\ref{sec:identifiability} shows that its normal gradient is then not
identified, which gives filtering a statistical rather than purely geometric
role, and the local guarantee
(Proposition~\ref{prop:local-geometric-ascent}) is stated for the true value
$Q^\pi$ under on-manifold agreement only.
\emph{Second}, RetraQ obtains both operators from a flow-matching policy rather
than a score or diffusion denoiser. Rescaling the interpolant as
$\mathbf x_t/t=\mathbf a+\sigma_t\mathbf z$ turns an action into a
Gaussian-corrupted observation at level $\sigma_{t_g}$, so one velocity
evaluation of the context-conditioned flow gives $R$ and one JVP gives
$J_R\mathbf g$ (Eq.~\eqref{eq:retraq-reference-map}), with no separate score
network, noise-schedule conversion, or per-context geometry model.
\emph{Third}, DRGD's guarantees assume a uniformly accurate denoiser and
Jacobian. For a learned flow, Proposition~\ref{prop:retraq-learned-geometry}
bounds how errors of the velocity field and of its input derivatives enter $R$
and $J_R$, and Section~\ref{sec:exp-geometry} measures how the guidance level
and action dimension trade tangent accuracy against normal suppression.
\emph{Fourth}, RetraQ embeds the update in policy improvement: refined actions are
stop-gradient targets for anchored distillation and offline-to-online learning
with a frozen geometry reference, the step is calibrated to a target
displacement, and fork critics supply action-gradient supervision on the
robot.

\section{Theoretical Results}
\label{app:geometric-refinement}

This appendix collects the results behind Sections~\ref{sec:identifiability}
and~\ref{sec:retraq-refinement}. Section~\ref{app:theory-identifiability} shows
what critic values on the behavior manifold determine;
Section~\ref{app:theory-improvement} shows that the ideal geometric update
improves the true action value; and Section~\ref{app:retraq-theory}, the main
result of this appendix, shows that the denoising map of a flow approximates
the two ideal operators. Standard facts from Riemannian optimization are cited
rather than re-derived~\citep{boumal}.

\paragraph{Setting.}

Fix a context $\mathbf s$ and an action $\mathbf a$ on a smooth embedded
manifold $\mathcal M=\mathcal M_{\mathbf s}\subset\mathbb R^D$.
We omit $\mathbf s$ from the critics.
All statements are local to $\mathbf a$ and use the Euclidean metric in
the normalized action coordinates of the flow.
Let $P_T(\mathbf a)$ and $P_N(\mathbf a)$ be the tangent and normal
orthogonal projectors.
Choose a sufficiently small neighborhood $U$ of $\mathbf a$ where the
nearest-point projection $\Pi=\Pi_{\mathbf s}$ is single-valued and smooth.
Then $\mathrm D\Pi(\mathbf a)=P_T(\mathbf a)$~\citep{boumal}.
Here $\mathrm D$ denotes the derivative of a map.
Figure~\ref{fig:critic-identifiability} illustrates the two geometric
issues discussed below.

\subsection{What Critic Values on the Manifold Determine}
\label{app:theory-identifiability}

\begin{proposition}[Tangential agreement and normal-gradient freedom]
\label{prop:identifiability-main}
\label{prop:gradient-identifiability}
Let $Q_1$ be continuously differentiable on $U$.
\begin{enumerate}
    \item[\textup{(i)}]
    If a continuously differentiable critic $Q_2$ agrees with $Q_1$ on
    $\mathcal M\cap U$, then
    \begin{equation}
        P_T(\mathbf a)\nabla Q_1(\mathbf a)
        =P_T(\mathbf a)\nabla Q_2(\mathbf a).
        \label{eq:tangent-identifiability}
    \end{equation}
    \item[\textup{(ii)}]
    For every $\mathbf n\in N_{\mathbf a}\mathcal M$, there exists a
    continuously differentiable $Q_2$ on $U$ that agrees with $Q_1$ on
    $\mathcal M\cap U$ and satisfies
    \begin{equation}
        \nabla Q_2(\mathbf a)=\nabla Q_1(\mathbf a)+\mathbf n.
        \label{eq:normal-gradient-freedom}
    \end{equation}
    Moreover, it can be chosen so that, for every $\mathbf x\in U$,
    \begin{equation}
        |Q_2(\mathbf x)-Q_1(\mathbf x)|
        \leq \|\mathbf n\|_2
              \operatorname{dist}(\mathbf x,\mathcal M).
        \label{eq:normal-perturbation-value-bound}
    \end{equation}
\end{enumerate}
\end{proposition}

\begin{proof}
For part~\textup{(i)}, take any
$\boldsymbol\xi\in T_{\mathbf a}\mathcal M$.
There is a smooth curve $\mathbf c$ on $\mathcal M$ with
$\mathbf c(0)=\mathbf a$ and $\mathbf c'(0)=\boldsymbol\xi$.
For a sufficiently short segment, this curve lies in $U$, where
$Q_2(\mathbf c(t))-Q_1(\mathbf c(t))=0$.
Differentiating at $t=0$ gives
\[
    \boldsymbol\xi^\top
    \bigl(\nabla Q_2(\mathbf a)-\nabla Q_1(\mathbf a)\bigr)=0.
\]
The gradient difference is orthogonal to every tangent vector, proving
Eq.~\eqref{eq:tangent-identifiability}.

For part~\textup{(ii)}, define
\[
    h_{\mathbf n}(\mathbf x)
    :=\mathbf n^\top\bigl(\mathbf x-\Pi(\mathbf x)\bigr),
    \qquad
    Q_2(\mathbf x):=Q_1(\mathbf x)+h_{\mathbf n}(\mathbf x).
\]
On $\mathcal M\cap U$, $\Pi(\mathbf x)=\mathbf x$, so $h_{\mathbf n}=0$.
At $\mathbf a$,
\[
    \nabla h_{\mathbf n}(\mathbf a)
    =\bigl(I_D-\mathrm D\Pi(\mathbf a)\bigr)^\top\mathbf n
    =\bigl(I_D-P_T(\mathbf a)\bigr)\mathbf n
    =\mathbf n.
\]
Finally, the Cauchy--Schwarz inequality and the nearest-point property give
\[
    |h_{\mathbf n}(\mathbf x)|
    \leq \|\mathbf n\|_2\,\|\mathbf x-\Pi(\mathbf x)\|_2
    =\|\mathbf n\|_2\operatorname{dist}(\mathbf x,\mathcal M).
\]
These identities establish both the gradient change and the value bound.
\end{proof}

The proposition assumes exact agreement throughout a local manifold
patch. It identifies the tangential derivative of that shared restriction,
without asserting that the shared values equal the true action value.
Agreement on finitely many samples, or a small value-fitting error, does
not by itself guarantee accurate tangential derivatives.

\paragraph{Data in a thin neighborhood.}
Let $\mathcal S_\delta\subset U$ be a set of data actions satisfying
$\operatorname{dist}(\mathbf x,\mathcal M)\leq\delta$ for every
$\mathbf x\in\mathcal S_\delta$, where $\delta>0$.
Let $\epsilon>0$ be an allowed discrepancy between the values predicted
by two critics on these actions.
For any $\mathbf n\in N_{\mathbf a}\mathcal M$ with
$\|\mathbf n\|_2\leq\epsilon/\delta$, the construction in
Proposition~\ref{prop:identifiability-main}\textup{(ii)} gives
\begin{equation}
    \sup_{\mathbf x\in\mathcal S_\delta}
    |Q_2(\mathbf x)-Q_1(\mathbf x)|
    \leq \delta\|\mathbf n\|_2
    \leq\epsilon,
    \qquad
    \nabla Q_2(\mathbf a)-\nabla Q_1(\mathbf a)=\mathbf n.
    \label{eq:thin-neighborhood-ambiguity}
\end{equation}
Thus, a normal-gradient change of size $\epsilon/\delta$ is compatible
with a value discrepancy of at most $\epsilon$ on all these data.
For a fixed value tolerance, this construction permits larger gradient
changes as the data become more concentrated around the manifold.

Here $\epsilon$ measures the difference between two critics, rather than
either critic's error against a target.
If $Q_1$ fits target values $y(\mathbf x)$ with uniform error at most
$\epsilon_0$ on $\mathcal S_\delta$, the triangle inequality gives
\[
    \sup_{\mathbf x\in\mathcal S_\delta}
    |Q_2(\mathbf x)-y(\mathbf x)|
    \leq\epsilon_0+\epsilon.
\]
In particular, if $Q_1$ fits these targets exactly, $Q_2$ fits them to
accuracy $\epsilon$ while allowing the stated normal-gradient change.

This is a constructive ambiguity result.
It does not upper-bound all possible normal-gradient errors by
$\epsilon/\delta$; such a bound would require additional assumptions
controlling derivatives.

\subsection{Local Improvement by the Geometric Update}
\label{app:theory-improvement}

Two standard facts connect the update of
Eq.~\eqref{eq:ideal-manifold-update} to Riemannian optimization under the
induced Euclidean metric~\citep{boumal}. First, $\mathbf g_T=P_T(\mathbf a)\nabla Q(\mathbf a)$
is the Riemannian gradient of the restriction $Q|_{\mathcal M}$, so the
normalized filtered gradient is the unit tangent direction of steepest
initial increase of $Q$ along $\mathcal M$. Second,
$\operatorname{Retr}_{\mathbf a}(\boldsymbol\xi):=\Pi(\mathbf a+\boldsymbol\xi)$
is a local retraction: $\operatorname{Retr}_{\mathbf a}(\mathbf 0)=\mathbf a$
and $\mathrm D\operatorname{Retr}_{\mathbf a}(\mathbf 0)[\boldsymbol\xi]=\boldsymbol\xi$
for tangent $\boldsymbol\xi$, because $\mathrm D\Pi(\mathbf a)=P_T(\mathbf a)$.
The following statement specifies when critic ascent along this retraction
also improves the true action value.

\begin{proposition}[Local action-value improvement]
\label{prop:local-geometric-ascent}
Let $\pi$ be a fixed continuation policy, and let $Q^\pi$ denote its true
action-value function at the fixed context.
Suppose $Q$ and $Q^\pi$ are continuously differentiable near
$\mathbf a\in\mathcal M$ and agree on a local manifold patch containing
$\mathbf a$.
For $\mathbf g_T=P_T(\mathbf a)\nabla Q(\mathbf a)$ and any local
retraction at $\mathbf a$,
\begin{equation}
    \left.
    \frac{\mathrm d}{\mathrm d\eta}
    Q^\pi\!\left(\operatorname{Retr}_{\mathbf a}(\eta\mathbf g_T)\right)
    \right|_{\eta=0}
    =\|\mathbf g_T\|_2^2.
    \label{eq:true-value-ascent}
\end{equation}
If $\mathbf g_T\ne\mathbf 0$, then
$Q^\pi(\operatorname{Retr}_{\mathbf a}(\eta\mathbf g_T))
>Q^\pi(\mathbf a)$ for all sufficiently small $\eta>0$.
\end{proposition}

\begin{proof}
Proposition~\ref{prop:identifiability-main}\textup{(i)} gives
$P_T(\mathbf a)\nabla Q^\pi(\mathbf a)=\mathbf g_T$. By the chain rule and
the retraction property,
\[
    \left.
    \frac{\mathrm d}{\mathrm d\eta}
    Q^\pi\!\left(\operatorname{Retr}_{\mathbf a}(\eta\mathbf g_T)\right)
    \right|_{\eta=0}
    =\nabla Q^\pi(\mathbf a)^\top\mathbf g_T
    =\|\mathbf g_T\|_2^2.
\]
A positive derivative at zero implies the stated strict increase for
all sufficiently small positive step sizes.
\end{proof}

Proposition~\ref{prop:local-geometric-ascent} assumes exact critic values on
a manifold patch and exact geometric operators, and concerns one context
under a fixed continuation policy. A learned filter can leave a residual
normal component, so the stepped action need not lie in the domain of a
retraction; its endpoint can still be corrected by $\Pi$ as long as it stays
in a tube where $\Pi$ is single-valued. Section~\ref{app:retraq-theory} shows how
the learned denoising map approximates both operators on such a tube.

\subsection{Flow-Based Approximation of the Geometric Operators}
\label{app:retraq-theory}

This section states and proves the two results that
Section~\ref{sec:retraq-refinement} uses to connect the denoising map
$R$ of Eq.~\eqref{eq:retraq-reference-map} to the ideal operators $P_T$ and
$\Pi_{\mathbf s}$ of Section~\ref{sec:identifiability}.
Proposition~\ref{prop:retraq-local-geometry} concerns the map
$R^{\mathrm{opt}}$ defined by the population-optimal velocity;
Proposition~\ref{prop:retraq-learned-geometry} quantifies the additional error
of a learned velocity.
As above, the context is fixed and omitted, and $\mu=\mu(\cdot\mid\mathbf s)$.
Norms of matrices are operator norms.

\paragraph{Assumptions.}
\begin{enumerate}
    \item[\textup{(A1)}] $\mathcal M\subset\mathbb R^D$ is a compact
    $k$-dimensional embedded $C^4$ submanifold without boundary, with
    positive reach $\tau_{\mathcal M}$~\citep{federer}.
    \item[\textup{(A2)}] $\mu$ is supported on $\mathcal M$ and has a density
    $p\in C^2(\mathcal M)$ with respect to the $k$-dimensional volume measure
    of $\mathcal M$, and $p>0$ on $\mathcal M$.
\end{enumerate}
Fix $0<r<\tau_{\mathcal M}$ and the closed tube
$U_r:=\{\mathbf a:\operatorname{dist}(\mathbf a,\mathcal M)\le r\}$.
On $U_r$, the nearest-point projection $\Pi$ is single-valued and
$C^{3}$~\citep{federer}, and its derivative at
$\mathbf a=\Pi(\mathbf a)+d\,\boldsymbol\nu$, with $d=\operatorname{dist}(\mathbf a,\mathcal M)$
and $\boldsymbol\nu$ a unit normal vector at $\Pi(\mathbf a)$, is
\begin{equation}
    \mathrm D\Pi(\mathbf a)
    =\bigl(I_T-d\,S_{\boldsymbol\nu}\bigr)^{-1}P_T\bigl(\Pi(\mathbf a)\bigr),
    \label{eq:dpi-tube}
\end{equation}
where $I_T$ is the identity on the tangent space at $\Pi(\mathbf a)$ and
$S_{\boldsymbol\nu}$ is the shape operator in direction $\boldsymbol\nu$,
the self-adjoint map on that space defined by
$\langle S_{\boldsymbol\nu}\boldsymbol\xi,\boldsymbol\zeta\rangle=\langle\mathrm{II}(\boldsymbol\xi,\boldsymbol\zeta),\boldsymbol\nu\rangle$,
with $\mathrm{II}$ the normal-valued second fundamental form; $I_T-dS_{\boldsymbol\nu}$ is
invertible because $d<\tau_{\mathcal M}$. On $\mathcal M$, $d=0$ and
$\mathrm D\Pi=P_T$~\citep{boumal}.

\paragraph{The ideal map is a Gaussian posterior mean.}
As in Section~\ref{sec:retraq-refinement}, write $\sigma=\sigma_g=(1-t_g)/t_g$,
and let $\mathbf A\sim\mu$ and $\mathbf Z\sim\mathcal N(0,I_D)$ be independent.
Then
$R^{\mathrm{opt}}(\mathbf a)=\mathbb E[\mathbf A\mid\mathbf A+\sigma\mathbf Z=\mathbf a]$,
and differentiating this Gaussian posterior mean in $\mathbf a$ gives
$J_R^{\mathrm{opt}}(\mathbf a)=\sigma^{-2}\operatorname{Cov}[\mathbf A\mid\mathbf A+\sigma\mathbf Z=\mathbf a]$
(second-order Tweedie identity~\citep{efron,manor2024posterior}).
The posterior of $\mathbf A$ given the observation $\mathbf a$ is the
probability measure on $\mathcal M$
\begin{equation}
    \rho_{\mathbf a}(\mathrm d\mathbf m)
    \;\propto\;
    p(\mathbf m)\exp\!\Bigl(-\frac{\|\mathbf a-\mathbf m\|_2^2}{2\sigma^2}\Bigr)\,
    \mathrm{vol}_{\mathcal M}(\mathrm d\mathbf m),
    \label{eq:manifold-posterior}
\end{equation}
so $R^{\mathrm{opt}}(\mathbf a)$ and $\sigma^2J_R^{\mathrm{opt}}(\mathbf a)$ are
the mean and covariance of $\rho_{\mathbf a}$.

\begin{proposition}[Local geometry of the ideal denoising map]
\label{prop:retraq-local-geometry}
Under \textup{(A1)--(A2)}, there exist $\sigma_0>0$ and $C<\infty$, depending
on $\mathcal M$, $p$, and $r$, such that for all $\sigma\in(0,\sigma_0]$ and all
$\mathbf a\in U_r$,
\begin{equation}
    \bigl\|R^{\mathrm{opt}}(\mathbf a)-\Pi(\mathbf a)\bigr\|_2\le C\sigma^2,
    \qquad
    \bigl\|J_R^{\mathrm{opt}}(\mathbf a)-\mathrm D\Pi(\mathbf a)\bigr\|\le C\sigma^2.
    \label{eq:retraq-local-geometry}
\end{equation}
In particular, $J_R^{\mathrm{opt}}(\mathbf a)=P_T(\mathbf a)+O(\sigma^2)$ for
$\mathbf a\in\mathcal M$, which is
Eq.~\eqref{eq:retraq-geometric-conditions}.
\end{proposition}

\begin{proof}
Fix $\mathbf a\in U_r$, let $\mathbf m_0=\Pi(\mathbf a)$, and write
$\mathbf a=\mathbf m_0+d\boldsymbol\nu$ as above.

\emph{Step 1: localization.}
Let $\phi(\mathbf m):=\tfrac12\|\mathbf a-\mathbf m\|_2^2-\tfrac12d^2\ge0$ on
$\mathcal M$. Because $\mathbf m_0$ is the unique minimizer of $\phi$ and
$U_r$ and $\mathcal M$ are compact, for every $\varepsilon>0$ there is
$\delta(\varepsilon)>0$, independent of $\mathbf a\in U_r$, such that
$\phi(\mathbf m)\ge\delta(\varepsilon)$ whenever
$\|\mathbf m-\mathbf m_0\|_2\ge\varepsilon$. The mass that
$\rho_{\mathbf a}$ assigns to this region is therefore at most
$c\,\sigma^{-k}e^{-\delta(\varepsilon)/\sigma^2}$, since the normalizing
constant of Eq.~\eqref{eq:manifold-posterior} is at least of order
$\sigma^k e^{-d^2/(2\sigma^2)}$. The same bound, multiplied by
$\operatorname{diam}(\mathcal M)^2$, controls the contribution of this region to
the mean and covariance. These terms are $o(\sigma^m)$ for every $m$ and can be
neglected.

\emph{Step 2: chart.}
By compactness of $\mathcal M$, $\varepsilon$ can be chosen, independently
of $\mathbf m_0$, so that $\mathcal M\cap\{\|\mathbf m-\mathbf m_0\|_2<\varepsilon\}$
is a graph over the tangent space at $\mathbf m_0$. We parametrize $\mathcal M$ near $\mathbf m_0$ as
$\mathbf m(\mathbf u)=\mathbf m_0+E\mathbf u+\tfrac12\mathrm{II}(\mathbf u,\mathbf u)+O(\|\mathbf u\|^3)$,
where $E\in\mathbb R^{D\times k}$ is an orthonormal basis of the tangent space
at $\mathbf m_0$, and $\mathbf u\in\mathbb R^k$. Then
\[
    \phi(\mathbf m(\mathbf u))
    =\tfrac12\mathbf u^\top H\mathbf u+\phi_3(\mathbf u)+O(\|\mathbf u\|^4),
    \qquad
    H=I_k-d\,E^\top S_{\boldsymbol\nu}E,
\]
with $\phi_3$ a cubic form; there is no linear term because $\mathbf a-\mathbf m_0$
is normal. $H$ is positive definite, uniformly over $U_r$, because
$d\le r<\tau_{\mathcal M}$. The density of $\mathbf u$ is proportional to
$q(\mathbf u)\exp(-\phi(\mathbf m(\mathbf u))/\sigma^2)$, where
$q(\mathbf u)=p(\mathbf m(\mathbf u))\sqrt{\det g(\mathbf u)}$ is $C^2$ and
positive, with $g$ the metric in the chart.

\emph{Step 3: Laplace expansion.}
Substitute $\mathbf u=\sigma\mathbf w$. The density of $\mathbf w$ is
proportional to
$\exp(-\tfrac12\mathbf w^\top H\mathbf w)\bigl(1+\sigma\,\ell(\mathbf w)+O(\sigma^2(1+\|\mathbf w\|^6))\bigr)$,
where $\ell(\mathbf w)=\nabla\log q(\mathbf 0)^\top\mathbf w-\phi_3(\mathbf w)$
is an odd polynomial. Every term of order $\sigma$ in the moments of
$\mathbf w$ is a Gaussian moment of odd total degree and vanishes. Hence
$\mathbb E[\mathbf w]=O(\sigma)$, $\mathbb E[\mathbf w\mathbf w^\top]=H^{-1}+O(\sigma^2)$,
and $\mathbb E[\mathbf w\otimes\mathbf w\otimes\mathbf w]=O(\sigma)$, uniformly
over $U_r$ because the constants depend continuously on $\mathbf a$ through
$H$, $\phi_3$, and $q$ on a compact set.

\emph{Step 4: mean and covariance.}
Returning to $\mathbf u=\sigma\mathbf w$,
$\mathbb E[\mathbf u]=O(\sigma^2)$ and
$\mathbb E[\|\mathbf u\|^2]=O(\sigma^2)$, so
\[
    R^{\mathrm{opt}}(\mathbf a)
    =\mathbf m_0+E\,\mathbb E[\mathbf u]
     +\tfrac12\mathbb E[\mathrm{II}(\mathbf u,\mathbf u)]+O(\sigma^3)
    =\Pi(\mathbf a)+O(\sigma^2).
\]
For the covariance,
$\operatorname{Cov}[\mathbf m]=E\operatorname{Cov}[\mathbf u]E^\top+O(\sigma^4)$:
the cross terms between $E\mathbf u$ and $\tfrac12\mathrm{II}(\mathbf u,\mathbf u)$
involve third moments of $\mathbf u$, which are $\sigma^3\cdot O(\sigma)$, and
the remaining terms are of fourth order in $\mathbf u$. Since
$\operatorname{Cov}[\mathbf u]=\sigma^2\bigl(H^{-1}+O(\sigma^2)\bigr)$,
\[
    J_R^{\mathrm{opt}}(\mathbf a)
    =\sigma^{-2}\operatorname{Cov}[\mathbf m]
    =EH^{-1}E^\top+O(\sigma^2).
\]
Finally, $EH^{-1}E^\top=(I_T-dS_{\boldsymbol\nu})^{-1}P_T(\mathbf m_0)=\mathrm D\Pi(\mathbf a)$
by Eq.~\eqref{eq:dpi-tube}. Both bounds are uniform over $U_r$ by Steps 1
and 3.
\end{proof}

\paragraph{What the proposition does and does not provide.}
Proposition~\ref{prop:retraq-local-geometry} is an asymptotic statement as
$\sigma\to0$ for a behavior distribution supported exactly on a manifold with
positive reach. Its constant $C$ grows with the curvature of $\mathcal M$, the
variation of $\log p$, and the proximity of $r$ to the reach; it gives no
numerical error at the operating point $\sigma_g\approx0.11$. The result does
not apply when two parts of the support lie within a distance comparable to
$\sigma$, for example nearby action modes, because the reach is then small.
The manifold is also assumed to have no boundary; near the boundary of a
manifold patch, the posterior is truncated on one side, and its mean is shifted
inward by $O(\sigma)$ rather than $O(\sigma^2)$.
Off the manifold, the Jacobian approximates $\mathrm D\Pi$, which rescales
tangent directions by $(I_T-dS_{\boldsymbol\nu})^{-1}$ and annihilates
normal ones; it equals $P_T$ only on $\mathcal M$.

\begin{proposition}[Error of the learned denoising map]
\label{prop:retraq-learned-geometry}
Let $v_{\bar\theta}$ be any velocity field that is differentiable in
$\mathbf x$, let $v^\star$ be the population-optimal velocity of
Section~\ref{sec:prelim}, and write
$\mathbf e(\mathbf x):=v_{\bar\theta}(\mathbf s,\mathbf x,t_g)-v^\star(\mathbf s,\mathbf x,t_g)$.
For every action $\mathbf a$,
\begin{align}
    R(\mathbf a)-R^{\mathrm{opt}}(\mathbf a)
    &=(1-t_g)\,\mathbf e(t_g\mathbf a),
    \label{eq:learned-map-error}\\
    J_R(\mathbf a)-J_R^{\mathrm{opt}}(\mathbf a)
    &=t_g(1-t_g)\,\partial_{\mathbf x}\mathbf e(t_g\mathbf a).
    \label{eq:learned-jacobian-error}
\end{align}
Consequently, under the assumptions of
Proposition~\ref{prop:retraq-local-geometry}, for $\mathbf a\in U_r$ and
$\sigma_g\le\sigma_0$,
\begin{equation}
    \|R(\mathbf a)-\Pi(\mathbf a)\|_2
    \le C\sigma_g^2+(1-t_g)\|\mathbf e(t_g\mathbf a)\|_2,
    \qquad
    \|J_R(\mathbf a)-\mathrm D\Pi(\mathbf a)\|
    \le C\sigma_g^2+t_g(1-t_g)\|\partial_{\mathbf x}\mathbf e(t_g\mathbf a)\|.
    \label{eq:learned-geometry-bound}
\end{equation}
Moreover, if $\|\partial_{\mathbf x}v_{\bar\theta}(\mathbf s,\mathbf x,t_g)\|\le L$
for all $\mathbf x$, then
$\|J_R(\mathbf a)-t_gI_D\|\le t_g(1-t_g)L$ for every $\mathbf a$.
\end{proposition}

\begin{proof}
By Eq.~\eqref{eq:retraq-reference-map},
$R(\mathbf a)=t_g\mathbf a+(1-t_g)v_{\bar\theta}(\mathbf s,t_g\mathbf a,t_g)$,
and $R^{\mathrm{opt}}$ is the same expression with $v^\star$. Subtracting gives
Eq.~\eqref{eq:learned-map-error}. Differentiating in $\mathbf a$, with the
factor $t_g$ from the input scaling
(Eq.~\eqref{eq:retraq-input-jacobian}), gives
Eq.~\eqref{eq:learned-jacobian-error}. The bounds of
Eq.~\eqref{eq:learned-geometry-bound} follow from the triangle inequality and
Proposition~\ref{prop:retraq-local-geometry}. For the last statement,
$J_R(\mathbf a)-t_gI_D=t_g(1-t_g)\,\partial_{\mathbf x}v_{\bar\theta}(\mathbf s,t_g\mathbf a,t_g)$.
\end{proof}

\paragraph{Consequences for the guidance level.}
Eq.~\eqref{eq:learned-jacobian-error} shows that the filter inherits the error
in the \emph{input derivative} of the velocity, not only the error in its
value. Flow-matching regression controls the value error in mean square under
the noisy marginal, but not its derivative, so a small
regression loss does not by itself make $J_R$ an accurate tangent filter. The normal gain measured in Section~\ref{sec:exp-geometry} is a
direct diagnostic of the combined error of Eq.~\eqref{eq:learned-geometry-bound}
on normal inputs. The last statement of Proposition~\ref{prop:retraq-learned-geometry}
formalizes the limit $t_g\to1$ discussed in
Section~\ref{sec:retraq-refinement}: for any velocity with input derivatives
bounded by $L$, $J_R\to I_D$, the unfiltered direction of raw critic ascent,
even though $J_R^{\mathrm{opt}}\to P_T$ on $\mathcal M$ by
Proposition~\ref{prop:retraq-local-geometry}. The two propositions thus
bound the two sides of the trade-off in the choice of $t_g$
(Appendix~\ref{app:retraq-guidance-level}): the $C\sigma_g^2$ term favors large
$t_g$, whereas a bounded learned derivative forces $J_R$ toward $t_gI_D$ as
$t_g\to1$.

\section{Implementation Details of RetraQ}
\label{app:retraq-implementation}

This appendix gives the details that Section~\ref{sec:method} omits: how the
operator is computed and calibrated (\ref{app:retraq-update-details}), the
critics (\ref{app:retraq-critic}), how the three settings use refined actions,
including the full offline-to-online algorithm
(\ref{app:retraq-policy}), and the hyperparameters (Table~\ref{tab:retraq-hparams}).

\paragraph{Conventions.}
Flow time follows Section~\ref{sec:prelim}, with $t=1$ at the data. The
$\pi_{0.5}$ code base uses the reversed time $t'=1-t$, so its configuration
value $t'_g=0.1$ is the same operating point as $t_g=0.9$ here
($\sigma_g\approx0.11$). \emph{Real robot} refers to phone-charging plug
insertion on an AgiBot G02 dual-arm robot with three cameras (head and both
wrists; Section~\ref{sec:exp-real}). Displacements are Euclidean norms of
whole action chunks in the flow's normalized action coordinates.

\subsection{The Operator in Practice}
\label{app:retraq-update-details}

\paragraph{Cost.}
We never form the $D\times D$ matrix $J_R$. One Jacobian--vector product
through $R$ at $\mathbf a$ with tangent $\mathbf g$, computed by forward-mode
automatic differentiation, returns both $R(\mathbf s,\mathbf a)$ and
$\widetilde{\mathbf g}=J_R\mathbf g$ at a small constant multiple of the cost
of one forward pass. An update therefore costs one critic gradient, one JVP,
and one further forward pass of $R$ at $\mathbf a+\eta\widetilde{\mathbf g}$.
We use $M=1$ refinement step on all platforms.

\paragraph{Guidance level.}
\phantomsection\label{app:retraq-guidance-level}%
For $0<t<1$ write
$R_t(\mathbf s,\mathbf a):=D_t^{\bar\theta}(\mathbf s,t\mathbf a)
=t\mathbf a+(1-t)v_{\bar\theta}(\mathbf s,t\mathbf a,t)$, so that $R=R_{t_g}$
in Eq.~\eqref{eq:retraq-reference-map}. Its action-input Jacobian is
\begin{equation}
    J_t(\mathbf s,\mathbf a)
    :=\partial_{\mathbf a}R_t(\mathbf s,\mathbf a)
    =tI_D+t(1-t)\,
      \partial_{\mathbf x}v_{\bar\theta}(\mathbf s,t\mathbf a,t).
    \label{eq:retraq-input-jacobian}
\end{equation}
The factor $t$ in the second term comes from the input scaling; the JVP
includes it. The level $t_g$ trades two errors
(Propositions~\ref{prop:retraq-local-geometry}
and~\ref{prop:retraq-learned-geometry}): a large $t_g$ makes the
$O(\sigma_g^2)$ terms small but drives a learned $J_R$ toward $t_gI_D$, and a
small $t_g$ lets the posterior average across curvature and nearby action
modes, which blurs both $R$ and $J_R$. We use $t_g=0.9$ on all platforms. This
point was fixed from operator diagnostics on OGBench and inherited by the
other platforms; it was not tuned on task success, and we ran no $t_g$ sweep
on the robot.

\paragraph{Step-size calibration.}
The scale of $\widetilde{\mathbf g}$ varies with the critic and the local
geometry, so we fix the displacement rather than $\eta$. On a calibration
batch $\{(\mathbf s_j,\mathbf a^{(0)}_j)\}_{j=1}^{B}$, with first-step
filtered gradients
$\widetilde{\mathbf g}_j=J_R(\mathbf s_j,\mathbf a^{(0)}_j)\nabla_{\mathbf a}\hat Q(\mathbf s_j,\mathbf a^{(0)}_j)$,
we solve
\begin{equation}
    \operatorname*{stat}_{j=1,\dots,B}
    \bigl\|R(\mathbf s_j,\mathbf a^{(0)}_j+\eta\widetilde{\mathbf g}_j)
          -R(\mathbf s_j,\mathbf a^{(0)}_j)\bigr\|_2
    =d_{\mathrm{target}}
    \label{eq:retraq-step-calibration}
\end{equation}
for $\eta$ by a log-scale grid search followed by bisection. The statistic is
the batch mean in offline-to-online learning and the median in offline
distillation and robot test-time guidance. $\eta$ is calibrated once and then
held fixed: on the first actor batch in offline-to-online learning, on a
held-out split of 2048 states in offline distillation, and on a bank of
fork-state samples for robot test-time guidance ($\eta=0.01$). A calibration
that does not converge is flagged, never clipped, and excluded from
displacement-matched comparisons.

Subtracting $R(\mathbf s_j,\mathbf a^{(0)}_j)$ removes the map's own offset
$R(\mathbf s,\mathbf a)-\mathbf a$, which is present even when
$\widetilde{\mathbf g}=\mathbf 0$, so $d_{\mathrm{target}}$ measures only
critic-induced movement. The offset is not small: on 135 robot fork states,
$\|R(\mathbf a)-\mathbf a\|_2$ has median $1.62$, about 65 times the deployed
$d_{\mathrm{target}}=0.025$. Multiplying $\hat Q$ by $\beta>0$ multiplies every
$\widetilde{\mathbf g}_j$ by $\beta$ and the calibrated $\eta$ by $1/\beta$, so
the calibrated step $\eta\widetilde{\mathbf g}$ does not depend on the critic's
scale. The controlled toy problem instead calibrates the total displacement
$\|R(\mathbf s,\mathbf a+\eta\widetilde{\mathbf g})-\mathbf a\|_2$.

\paragraph{Padded actions.}
For padded action representations (on the robot, 16 executable coordinates
padded to 32, and 30 executed frames padded to the 50-frame horizon), the
fixed orthogonal projector $P_{\mathrm{act}}$ retains executable coordinates
and zeroes padded ones. The effective critic is
$\widehat Q(\mathbf s,\mathbf a):=\hat Q(\mathbf s,P_{\mathrm{act}}\mathbf a)$,
and step~\stepA{} of Eq.~\eqref{eq:retraq-step} uses its gradient. All other
operations and norms use model action coordinates.

\paragraph{Guards.}
\emph{Numerical} checks stop the run if a non-finite value appears in the
critic value, its gradient, the JVP output, the proposal, the refined action,
or $R(\mathbf a)$, or if $\eta$ leaves $[10^{-6},10^{2}]$; values are never
patched, since a patched target would train the actor toward an arbitrary
action without a trace. A \emph{displacement} check discards an update if the
critic-induced displacement exceeds a cap ($0.5$ by default, $0.25$ on the
robot) rather than rescaling it, because rescaling would still train on a
displacement at which neither the retraction nor the critic's local
linearization is trusted. The cap matters because matching a batch statistic
does not bound each action: on 240 robot fork frames, $\|\nabla_{\mathbf a}Q\|$
has median $0.0155$ and maximum $2.03$, and $2.5\%$ of the calibrated steps
exceed the robot cap.

\subsection{Critics}
\label{app:retraq-critic}

RetraQ uses only the action gradient of a critic. Table~\ref{tab:retraq-critics}
lists the critic used in each setting.

\begin{table}[t]
\centering
\caption{\textbf{Critics used by RetraQ.} TTG: test-time guidance; MC: Monte
Carlo; TD: temporal difference. FQL-style twin-$Q$ TD evaluates the bootstrap
target at an action of the current actor.}
\label{tab:retraq-critics}
\small
\setlength{\tabcolsep}{4pt}
\begin{tabular}{@{}p{0.27\linewidth}p{0.36\linewidth}p{0.33\linewidth}@{}}
\toprule
Platform and setting & Critic & Training data and targets\\
\midrule
Real robot, TTG & fork critic: Critic~1 (linear), Critic~2 (nonlinear) & 270 forked rollouts; binary success\\
Real robot, offline distill. & fork critic (Critic~1) & same\\
Real robot, offline-to-online & IQL, twin $Q$ and $V$, Eq.~\eqref{eq:retraq-iql} & online replay; terminal success\\
LIBERO, offline & chunk critic, MC & forked BC rollouts; discounted success\\
LIBERO, online & FQL-style twin-$Q$ TD, warm-started from MC & mixed replay; discounted success\\
RoboMimic Can, offline & chunk critic, MC & forked BC rollouts; discounted success\\
OGBench, offline and online & FQL-style twin-$Q$ TD & offline data, then mixed replay; benchmark reward\\
Toy & analytic $\hat Q$ (no learned critic) & ---\\
\bottomrule
\end{tabular}
\end{table}

\subsubsection{Fork Critic for Test-Time Guidance}
\label{app:retraq-data}

\begin{figure}[t]
    \centering
    \includegraphics[width=\textwidth]{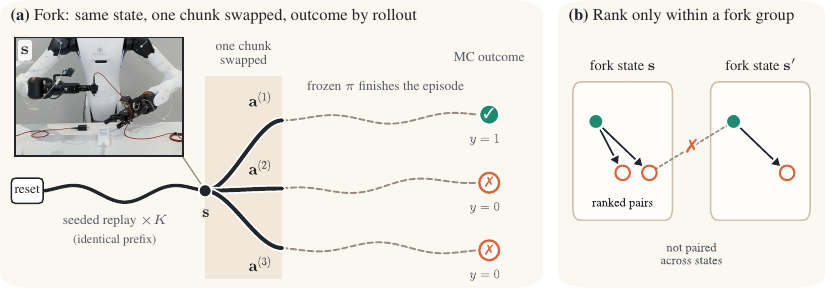}
    \caption{
    \textbf{Forked rollouts provide local action comparisons.}
    \textbf{(a)}~Seeded replay from reset reproduces the same prefix $K$
    times and reaches the fork state $\mathbf s$. In each branch, one action
    chunk is swapped for a perturbed chunk, the frozen policy $\pi_{\bar\theta}$
    finishes the episode, and the branch is labeled by its Monte Carlo task
    outcome $y$.
    \textbf{(b)}~Ranked pairs (arrows) are formed only within a fork group and
    place a successful branch (filled) above a failed one (hollow). Branches
    from different fork states $\mathbf s$ and $\mathbf s'$ are never paired,
    because their outcome difference would mix the effect of the state with
    that of the action.
    }
    \label{fig:fork-mechanism}
\end{figure}

\paragraph{Collection.}
A fork group is collected as in Figure~\ref{fig:fork-mechanism}a. Seeded
replay reproduces the same prefix and reaches the fork state $\mathbf s$. In
each branch, the first action chunk after $\mathbf s$ is replaced, the frozen
policy $\pi_{\bar\theta}$ finishes the episode, and the branch is labeled by
its task outcome $y$. The reference chunk $\mathbf a_c^{\mathrm{ref}}$ is the
chunk that the frozen policy predicts at the fork state. Because branches share
the state and differ only in the replaced chunk, outcome differences within a
group can be attributed to the action.

\paragraph{Robot fork states and branches.}
The robot fork states are taken at two task-critical moments: the right arm has
grasped the plug and is about to move it down toward the socket, and the plug
is very close to the socket. Each collection run resets the robot to a fixed
joint configuration at one of these moments. Runs are assigned to fork states
by clustering their frame-0 joint proprioception; five collection runs yield
four fork states. Scene variables that proprioception does not capture, such
as the pose of the phone and cable, are not matched, so branches are not exact
counterfactuals. A robot fork group contains three kinds of branches
(Table~\ref{tab:fork-tiers}). \emph{Injected-noise} branches add a zero-mean
Gaussian offset, constant over the chunk, to the coordinates
$\mathcal J=\{7,8,9\}$, joints 1--3 of the right arm that performs the
insertion, with standard deviation $\sigma_{\mathrm b}=0.3$ times the
per-coordinate action standard deviation. \emph{Policy-resample} branches
execute a new sample of the frozen policy, and \emph{human-teleoperated}
branches an operator-provided first chunk. The first chunk has
$H_{\mathrm{exec}}=30$ frames (1\,s at 30\,Hz). Of 300 collected rollouts,
270 are usable ($n=103,105,30,32$ over the four fork states; overall success
$34.1\%$): we exclude rollouts with an operator intervention at any frame,
whose later actions are not generated by the policy, and two whose first
chunk is incomplete (24 and 27 frames).

\begin{table}[t]
\centering
\caption{\textbf{Robot fork corpus by branch type.} Deviation is the
Euclidean norm of the first-chunk deviation from the reference chunk on the
perturbed joints $\mathcal J$, in radians. The tier weight enters the
critic loss, Eq.~\eqref{eq:fork-loss-exact}.}
\label{tab:fork-tiers}
\small
\begin{tabular}{lccccc}
\toprule
Branch type & $n$ & Success & Deviation, median & Deviation, max & Tier weight\\
\midrule
Injected noise    & 135 & 0.348 & 0.107 & 0.320 & 1.0\\
Human teleoperation & 32 & 0.469 & 0.035 & 0.084 & 0.5\\
Policy resample   & 103 & 0.291 & 0.024 & 0.094 & 0.25\\
\bottomrule
\end{tabular}
\end{table}

\paragraph{Simulation forks.}
In simulation, forks are taken along replayed BC trajectories: every 12
environment steps on LIBERO, and at steps 48, 96, and 144 on RoboMimic Can.
LIBERO branches perturb the chunk with independent Gaussian noise, and
RoboMimic branches with temporally correlated AR(1) noise (coefficient $0.7$,
rescaled to unit standard deviation). Simulation labels are discounted Monte
Carlo success $\gamma^T$, with $T$ the number of steps to success; robot labels
are binary terminal success judged by the operator.

\paragraph{Action input and heads.}
Let $\boldsymbol\Delta=\mathbf a-\mathbf a_c^{\mathrm{ref}}$ be the deviation in
Eq.~\eqref{eq:fork-critic}, $\boldsymbol\Delta_h$ its $h$-th control step,
and $P_{\mathcal J}$ the selector of the perturbed coordinates. The action
term reads the deviation through its mean over the executed chunk,
\begin{equation}
    \mathbf u
    =\bar P\boldsymbol\Delta
    :=\frac{1}{H_{\mathrm{exec}}}\sum_{h=1}^{H_{\mathrm{exec}}}
     P_{\mathcal J}\boldsymbol\Delta_h
    \in\mathbb R^{3},
    \qquad
    A_\omega(\boldsymbol\Delta)=f_\omega(\bar P\boldsymbol\Delta),
    \quad f_\omega(\mathbf 0)=0,
    \label{eq:fork-action-deviation}
\end{equation}
which matches how the injected perturbation is applied. Because $\bar P$ is
fixed and linear,
$\nabla_{\mathbf a}Q_\phi=\kappa\,\bar P^{\top}\nabla f_\omega(\mathbf u)$ is zero
on unperturbed and padded coordinates and beyond the executed chunk. The
baseline is linear in the 16-dimensional joint proprioception,
$B_\psi(\mathbf s)=\boldsymbol\psi^{\top}[\mathbf p(\mathbf s);1]$, and carries
almost no outcome information: alone, its held-out AUROC on the four fork
states is $0.511$, $0.527$, $0.578$, and $0.410$, whereas the action term
reaches $0.727$, $0.685$, $0.814$, and $0.508$. The robot uses two action
heads. \emph{Critic~1} is linear, $f_\omega(\mathbf u)=\mathbf w^\top\mathbf u$:
\begin{equation}
    Q_\phi(\mathbf s,\mathbf a)
    =\operatorname{sigmoid}\bigl(B_\psi(\mathbf s)
      +\mathbf w^{\top}\bar P(\mathbf a-\mathbf a^{\mathrm{ref}}_{c(\mathbf s)})\bigr),
    \qquad \phi=(\psi,\mathbf w).
    \label{eq:fork-critic-linear}
\end{equation}
\emph{Critic~2} is nonlinear in the deviation,
$f_\omega(\mathbf u)=\mathrm{MLP}_\omega(\mathbf u)-\mathrm{MLP}_\omega(\mathbf 0)$,
and is trained on the same data. Fork states whose action standard deviation
on $\mathcal J$ is below $0.25\sigma_{\mathrm b}=0.075$ do not train the
action head; this excludes state~3 ($0.052$).

\paragraph{Loss.}
The fork critic is trained with
\begin{equation}
    \mathcal L(\phi)
    =\mathcal L_{\mathrm{out}}(\phi)
    +\lambda_{\mathrm{rank}}\mathcal L_{\mathrm{rank}}(\omega),
    \qquad \lambda_{\mathrm{rank}}\geq0 .
    \label{eq:fork-loss}
\end{equation}
The outcome loss fits the predictions to the labels. The ranking loss compares
branches of the same fork group and asks the one with the better outcome to
receive the higher action score; since both branches share $\mathbf s$, the
baseline cancels and $\mathcal L_{\mathrm{rank}}$ depends on $\omega$ alone.
Pairs across fork groups are excluded (Figure~\ref{fig:fork-mechanism}b),
because their outcome difference would also reflect the state. With logits
$\nu_i=B_\psi(\mathbf s_i)+f_\omega(\mathbf u_i)$, tier weights $\alpha_i$
(Table~\ref{tab:fork-tiers}), and ranked pairs
$\mathcal P=\{(i,j):c_i=c_j,\ y_i>y_j+\epsilon\}$ with $\epsilon=0.02$, which
drops pairs with nearly equal labels,
\begin{equation}
    \mathcal L(\phi)
    =\lambda_{\mathrm{out}}\frac{\sum_i\alpha_i\,\mathrm{BCE}(\nu_i,y_i)}{\sum_i\alpha_i}
    +\lambda_{\mathrm{rank}}\frac{\sum_{(i,j)\in\mathcal P}\alpha_{ij}\,
      \log\bigl(1+e^{-(\nu_i-\nu_j)}\bigr)}{\sum_{(i,j)\in\mathcal P}\alpha_{ij}}
    +\lambda_{\mathbf w}\,\mathbb E_i\bigl\|\nabla_{\mathbf u}f_\omega(\mathbf u_i)\bigr\|_2^2,
    \label{eq:fork-loss-exact}
\end{equation}
where $\alpha_{ij}=\min(\alpha_i,\alpha_j)$, $\lambda_{\mathrm{out}}=0.1$,
$\lambda_{\mathrm{rank}}=1$, and $\lambda_{\mathbf w}=10^{-2}$. For the linear
head, the last term is $\|\mathbf w\|_2^2$. We train with learning
rate $3\times10^{-3}$ for 2000 steps.

\begin{figure}[t]
\centering
\includegraphics[width=0.56\linewidth]{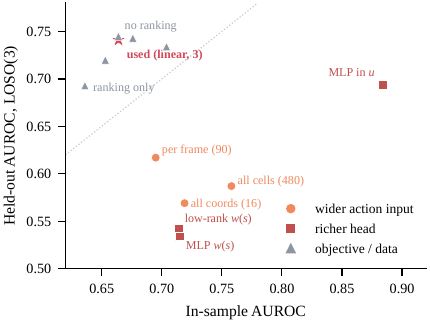}
\caption{\textbf{Fit versus transfer of fork-critic variants.} In-sample
AUROC versus leave-one-state-out AUROC over fork states 0--2. Wider action
inputs and richer heads move right (better fit) and down (worse transfer);
objective and data choices leave the used linear head's transfer nearly
unchanged. ``MLP in $\mathbf u$'' is Critic~2. Dotted line: $y=x$.}
\label{fig:fork-ablation}
\end{figure}

\paragraph{What the ablation shows.}
With four fork states and 270 rollouts, only a low-dimensional action
direction can be resolved (Figure~\ref{fig:fork-ablation}). Every richer
action input or head raises in-sample AUROC but lowers the AUROC on held-out
fork states: the linear head on the three-dimensional chunk mean reaches
$0.741$ in leave-one-state-out evaluation over states 0--2, whereas per-frame,
all-coordinate, and state-dependent variants fall to $0.534$--$0.617$, and the
nonlinear head of Critic~2 reaches $0.694$ despite the highest in-sample AUROC
($0.884$). This is why Critic~1 is the default. Critic~2 failed a gate fixed
in advance (held-out AUROC above $0.667$ when state~3 is also held out; it
reaches $0.649$), but it is evaluated as a second test-time critic
(Section~\ref{sec:exp-real}) because, unlike Critic~1, its gradient varies with
the action. Objective and data choices barely change transfer; in particular,
removing the ranking term leaves the held-out AUROC at $0.745$, so the fork
critic's transfer does not depend on it. Across three training seeds, the
cosine similarity of the linear head's $\nabla_{\mathbf a}Q$ exceeds $0.9999$.

\paragraph{What a linear head implies for RetraQ.}
With Eq.~\eqref{eq:fork-critic-linear}, the action gradient is
$\kappa\,\bar P^{\top}\mathbf w$. Its direction is the same at every action of
a fork group, and $Q_\phi$ increases monotonically along it with no interior
maximum. The critic therefore says which way to move but not how far: the
step length comes from the displacement calibration of
Eq.~\eqref{eq:retraq-step-calibration}, not from a critic maximum. The data
support this direction only within the perturbation range, while a linear
head extrapolates beyond it. The deployed step stays well inside that range: a
median step changes each coordinate by at most $d_{\mathrm{target}}=0.025$,
whereas the injected perturbations have standard deviation
$\sigma_{\mathrm b}=0.3$ on each perturbed coordinate.

\subsubsection{Replay Critics for Offline-to-Online Learning}
\label{app:retraq-replay-critic}

Online rollouts are not forked, so no two samples share a state and the
ranking term is unavailable. A replay transition stores a context $\mathbf s$,
the executed part of a chunk ($k\le H$ frames, padded to the horizon $H$ with
a frame mask $\mathbf m$; Appendix~\ref{app:retraq-policy}), the per-frame
rewards $r_0,\dots,r_{k-1}$, the next context $\mathbf s'$, and a terminal
flag $d$. On the robot, we train twin critics $Q_{\phi_1},Q_{\phi_2}$ and a
state value $V_\psi$ with implicit Q-learning~\citep{kostrikov2022iql}, with
target critics $Q_{\bar\phi_i}$:
\begin{equation}
    \begin{aligned}
    \mathcal L_V(\psi)&=\mathbb E\bigl[L_2^{\tau}\bigl(\bar q-V_\psi(\mathbf s)\bigr)\bigr],
    &\bar q&=\operatorname{sg}\bigl[\min_{i}Q_{\bar\phi_i}(\mathbf s,\mathbf a,\mathbf m)\bigr],\\
    \mathcal L_Q(\phi_1,\phi_2)&=\mathbb E\Bigl[\textstyle\sum_{i=1}^{2}\bigl(Q_{\phi_i}(\mathbf s,\mathbf a,\mathbf m)-y\bigr)^2\Bigr],
    &y&=\operatorname{sg}\Bigl[\textstyle\sum_{h=0}^{k-1}\gamma^h r_h+\gamma^{k}(1-d)\,V_\psi(\mathbf s')\Bigr],
    \end{aligned}
    \label{eq:retraq-iql}
\end{equation}
where $L_2^{\tau}(x)=|\tau-\mathbb 1(x<0)|\,x^2$ is the expectile loss. We use
$\tau=0.7$, $\gamma=0.99$ per executed frame, and Polyak averaging
$\bar\phi_i\gets(1-\rho)\bar\phi_i+\rho\phi_i$ with $\rho=0.005$ after every
update. The reward is sparse: $r=1$ at the terminal frame of a success judged
by the operator and $0$ otherwise. The networks read frozen
vision--language features of the BC model (masked mean pooling, concatenated
with the normalized robot state; 2080 dimensions). $Q$ additionally takes the
$30\times16$ executed actions and the 30-dimensional frame mask (2590 inputs
in total); $Q$ and $V$ are MLPs with two hidden layers of width 256, LayerNorm,
and tanh, with no output activation, and share no trainable parameters. Each
is trained with Adam, learning rate $3\times10^{-4}$, and gradient-norm
clipping at $1.0$, in FP32. RetraQ reads the frozen snapshot
$\hat Q=\min_iQ_{\phi_i}$. IQL evaluates the critics only at executed actions,
so their values are supported by data while their normal gradients are
unconstrained, which is the case step~\stepB{} handles.

In simulation, LIBERO and OGBench use FQL-style twin-$Q$ TD critics, whose
bootstrap target evaluates the next action of the current actor. The LIBERO
online critic is warm-started from the offline Monte Carlo chunk critic
(8000 training steps); the OGBench critic is first trained for $10^6$ offline
steps, frozen during offline RetraQ, and then trained online. Both mix offline
and online data equally in critic batches.

\subsection{The Three Settings}
\label{app:retraq-policy}

Table~\ref{tab:retraq-settings} summarizes how the settings of
Section~\ref{sec:retraq-deployment} use the shared operator \textsc{Refine}.

\begin{table}[t]
\centering
\caption{\textbf{The three RetraQ settings} share the operator \textsc{Refine}
(Algorithm~\ref{alg:retraq-training}) and differ in its inputs and in how its
output is used. Critics per platform are listed in Table~\ref{tab:retraq-critics}.}
\label{tab:retraq-settings}
\small
\begin{tabular}{@{}llll@{}}
\toprule
Setting & Initial action $\mathbf a^{(0)}$ & Critic & Use of $\mathbf a^\star$\\
\midrule
Test-time guidance & $\pi_{\bar\theta}(\cdot\mid\mathbf s)$ at fork states & fork critic, Eq.~\eqref{eq:fork-critic} & executed\\
Offline distillation & current actor at dataset states & fixed during refinement & distilled\\
Offline-to-online & executed chunks in replay $\mathcal D$ & replay critic, snapshot $\hat Q$ & distilled\\
\bottomrule
\end{tabular}
\end{table}

\paragraph{Test-time guidance on the robot.}
With all parameters fixed, RetraQ refines a sample of the BC policy before
execution, only at detected fork states. The detector computes the Euclidean
distance from the 16-dimensional joint proprioception to the four stored
fork-state centroids and selects the nearest one if the distance is below
$0.15$ (within-state spread at most $0.007$\,rad; centroids at least $0.38$
apart). Guidance is restricted to states~0--2, since state~3 lies below the
action-standard-deviation floor and its held-out AUROC is $0.446$. Otherwise,
and if the critic logit saturates ($|\nu|>15$) or a guard fails, the BC chunk
is executed unchanged. Critic~1 and Critic~2 use the same detector, operator,
and step size.

\paragraph{Offline distillation.}
Offline distillation runs outer iterations on a fixed dataset without
collection; the critic is not updated while targets are refined. In each iteration, the current actor samples an
action at each dataset state (detached), \textsc{Refine} improves it, and the
actor is distilled toward the refined actions. Each iteration fits a new set of
targets, so the actor's Adam optimizer is re-initialized at its start. We run
30 iterations on the toy and 5 on LIBERO, RoboMimic, and OGBench, and report
intermediate iterations where stated (LIBERO $k\le3$, RoboMimic $k=2$). On
OGBench, offline RetraQ uses no anchor term ($\lambda_{\mathrm{BC}}=0$). On the
robot, the offline-distillation arm instead fine-tunes $\pi_{0.5}$
continuously on refined targets from the Critic~1 fork critic, with the
deployed guidance level $t_g=0.9$ ($t'_g=0.1$ in the $\pi_{0.5}$ convention)
and a total-displacement target of $2.10$ instead of the test-time
$d_{\mathrm{target}}=0.025$.

\paragraph{Offline-to-online learning.}
Refinement starts from replayed actions. A replayed transition contains the
executed part of a chunk, $\mathbf a^{\mathrm{exec}}$, with $k\le H$ frames. We
convert it to the normalized action coordinates of the BC flow (on the robot,
absolute to delta actions, then normalization), zero-pad it to the horizon,
$\mathbf a^{(0)}=\mathrm{Pad}_H(\mathbf a^{\mathrm{exec}},\mathbf 0)$, and mark
the executed frames with $m_h=\mathbb 1[h\le k]$. The critic takes the mask as
input, and the improvement loss is restricted to the executed frames,
\begin{equation}
    \mathcal L_{\mathrm{FM}}(\theta;\mathbf s,\mathbf a,\mathbf m)
    :=\mathbb E_{t,\mathbf z}\bigl\|\mathbf m\odot\bigl(v_\theta(\mathbf s,\mathbf x_t,t)-(\mathbf a-\mathbf z)\bigr)\bigr\|_2^2,
    \label{eq:retraq-masked-fm}
\end{equation}
where $\odot$ multiplies each frame by its mask entry; the loss of
Section~\ref{sec:retraq-deployment} is the case $\mathbf m=\mathbf 1$. The actor
loss is
$\mathbb E_{\mathcal B_\pi}\bigl[\lambda_{\mathrm{imp}}\mathcal L_{\mathrm{FM}}(\theta;\mathbf s,\mathbf a^\star,\mathbf m)
+\lambda_{\mathrm{BC}}\mathcal L_{\mathrm{FM}}(\theta;\mathbf s,\mathbf a^{\mathrm{BC}},\mathbf 1)\bigr]$,
where $\mathbf a^{\mathrm{BC}}$ is a freshly generated full-horizon chunk of
the frozen BC policy on every platform. The improvement term supervises only
the executed frames; the anchor term supplies targets for every frame.
Gradients reach only $\theta$, never target generation or the sampling chain.
The BC flow $\bar\theta$ stays frozen throughout.
Algorithm~\ref{alg:retraq-online-full} gives the robot version. Updates are
triggered by new data: on the robot, every 20 new chunk transitions allow one
actor update attempt and two critic updates, each drawing a fresh global batch
of 128 chunk transitions (16 per GPU on 8 GPUs) entirely from online replay. An
attempt whose guards fail is skipped, so the number of actor updates can be
smaller than the number of attempts. No human interventions occurred in the reported
online runs. In simulation, updates proceed in blocks (LIBERO: 1500 critic
steps per 2500 environment steps; OGBench: 5000 critic steps per 5000
environment steps), with $\eta$ calibrated once as on the robot.

\begin{algorithm}[t]
\caption{Offline-to-online RetraQ on the robot, full version of
Algorithm~\ref{alg:retraq-training}, with frame masks, guards, and the IQL
critic update of Eq.~\eqref{eq:retraq-iql}. \textsc{Refine} is defined in
Algorithm~\ref{alg:retraq-training}; here $\hat Q$ also takes the mask
$\mathbf m$.}
\label{alg:retraq-online-full}
\small
\begin{algorithmic}[1]
\Require BC parameters $\bar\theta$; demonstrations $\mathcal D_{\mathrm{demo}}$; replay buffer $\mathcal D$;
         horizon $H$; refinement steps $M$; guidance level $t_g$; target displacement $d_{\mathrm{target}}$;
         IQL parameters $\gamma,\tau,\rho$; loss weights $\lambda_{\mathrm{imp}},\lambda_{\mathrm{BC}}$;
         warm-up updates $W$; critic updates per cycle $K_Q$
\State $\theta\gets\bar\theta$ ($\bar\theta$ frozen); initialize $Q_{\phi_1},Q_{\phi_2},V_\psi$; $\bar\phi_i\gets\phi_i$; $\eta\gets$ unset
\For{$w=1,\dots,W$}
    \State \Call{IQLUpdate}{batch from $\mathcal D_{\mathrm{demo}}$}
    \Comment{actor fixed}
\EndFor
\While{training budget remains}
    \State Collect rollouts of $\pi_\theta$; label success or failure; append transitions to $\mathcal D$
    \For{each update cycle allowed by the replay schedule}
        \For{$j=1,\dots,K_Q$}
            \State \Call{IQLUpdate}{batch from $\mathcal D$}
        \EndFor
        \State Freeze $\hat Q(\mathbf s,\mathbf a,\mathbf m)=\min_iQ_{\phi_i}(\mathbf s,\mathbf a,\mathbf m)$
        \State Sample actor batch $\mathcal B_\pi$ of $(\mathbf s,\mathbf a^{\mathrm{exec}},k)$ from $\mathcal D$
        \State $\mathbf a^{(0)}\gets\mathrm{Pad}_H(\mathbf a^{\mathrm{exec}},\mathbf 0)$;\;
               $m_h\gets\mathbb 1[h\le k]$
               \Comment{normalized coordinates}
        \State \textbf{if} $\eta$ is unset \textbf{then} solve Eq.~\eqref{eq:retraq-step-calibration} on $\mathcal B_\pi$
               \Comment{then held fixed}
        \State $\mathbf a^\star\gets\Call{Refine}{\mathbf s,\mathbf a^{(0)};\,\hat Q,\eta}$;\;
               $\mathbf a^{\mathrm{BC}}\sim\pi_{\bar\theta}(\cdot\mid\mathbf s)$
               \Comment{full horizon}
        \State $\mathcal L_\pi\gets\mathbb E_{\mathcal B_\pi}\bigl[\lambda_{\mathrm{imp}}\mathcal L_{\mathrm{FM}}(\theta;\mathbf s,\mathbf a^\star,\mathbf m)
               +\lambda_{\mathrm{BC}}\mathcal L_{\mathrm{FM}}(\theta;\mathbf s,\mathbf a^{\mathrm{BC}},\mathbf 1)\bigr]$
        \State Update $\theta$ with $\nabla_\theta\mathcal L_\pi$ if the numerical and displacement guards pass
    \EndFor
    \State Periodically synchronize actor weights to the rollout workers
\EndWhile
\Statex
\Procedure{IQLUpdate}{$\mathcal B$}
    \State Update $V_\psi$ on $\mathcal L_V$, then $Q_{\phi_1},Q_{\phi_2}$ on $\mathcal L_Q$ (Eq.~\eqref{eq:retraq-iql})
    \State $\bar\phi_i\gets(1-\rho)\bar\phi_i+\rho\phi_i$ for $i=1,2$
\EndProcedure
\end{algorithmic}
\end{algorithm}

\subsection{Hyperparameters}

Table~\ref{tab:retraq-hparams} lists the RetraQ hyperparameters on each platform.

\begin{table}[t]
\centering
\caption{\textbf{RetraQ hyperparameters.} Values shared by all methods on a
platform (policy backbone, evaluation protocol) are in
Appendix~\ref{app:retraq-experiments}; critics are listed in
Table~\ref{tab:retraq-critics}.}
\label{tab:retraq-hparams}
\small
\setlength{\tabcolsep}{3.2pt}
\begin{tabular}{lccccc}
\toprule
 & Toy & LIBERO & RoboMimic & OGBench & Real robot \\
\midrule
Guidance level $t_g$ ($\sigma_g$) & \multicolumn{4}{c}{0.9 (0.11)} & 0.9 (0.11)$^{\ddagger}$ \\
Target displacement $d_{\mathrm{target}}$ & 0.05$^{\dagger}$ & 0.25 & 0.10 & 0.05 & 0.025 \\
Refinement steps $M$ & \multicolumn{5}{c}{1} \\
Displacement cap & \multicolumn{4}{c}{0.5} & 0.25 \\
Executed frames per chunk & -- & 5 & 8 & 1 & 30 \\
Flow integration steps & 100 & 32 & 64 & 128 & 10 \\
Offline outer iterations & 30 & 5 & 5 & 5 & -- \\
Critic steps before improvement & -- & 8{,}000 & 12{,}000 / iter. & $10^6$ & see$^{\S}$ \\
Critic steps per online block & -- & 1{,}500 & -- & 5{,}000 & 2 / 20 trans. \\
Actor batch size & -- & 256 & 256 & 512 & 128 \\
Loss weights $\lambda_{\mathrm{imp}}$ / $\lambda_{\mathrm{BC}}$ & -- & 1 / 1 & 1 / 1 & 1 / 0 (offline) & 1 / 1 \\
\bottomrule
\end{tabular}

\vspace{2pt}
{\footnotesize $^{\dagger}$The toy calibrates the total displacement
$\|R(\mathbf s,\mathbf a+\eta\widetilde{\mathbf g})-\mathbf a\|_2$; all other
platforms calibrate the critic-induced displacement of
Eq.~\eqref{eq:retraq-step-calibration}.
$^{\ddagger}$Configured as $t'_g=0.1$ in the reversed time convention of the
$\pi_{0.5}$ code base.
$^{\S}$Test-time guidance: fork critic, 2000 steps on 270 forked rollouts.
Offline-to-online: IQL critic warmed up on demonstrations, then 2 updates per
20 new transitions.
On the robot, the actor is initialized from a $\pi_{0.5}$ BC checkpoint
trained for 80{,}000 steps on the task demonstrations and trained with AdamW
($\beta=(0.9,0.95)$, learning rate $3.75\times10^{-5}$, 1000-step linear
warm-up, cosine decay to $7.5\times10^{-6}$) with the vision encoder frozen.}
\end{table}   

\section{Additional Experimental Details and Results}
\label{app:retraq-experiments}

\subsection{Operator Definitions and Calibration}

We use the abbreviations $R$, $J$, and $\mathbf g$ introduced in
Section~\ref{sec:exp-setup}.
Table~\ref{tab:retraq-operator-definitions} specifies the target used by
each component ablation.
The unchanged-target control measures changes caused by repeated
self-distillation. The map-only control measures the result of applying
the frozen action map without a critic step.
Neither control is displacement-matched to the critic-guided arms.

\begin{table}[t]
\centering
\small
\caption{Targets for component ablations. The step size $\eta$ is
calibrated separately for each guided arm.}
\label{tab:retraq-operator-definitions}
\begin{tabular}{@{}ll@{}}
\toprule
Name & Target $\mathbf a^\star$\\
\midrule
Self-distillation & $\mathbf a$\\
Map-only & $R(\mathbf a)$\\
Raw ascent & $\mathbf a+\eta\mathbf g$\\
Filter-only & $\mathbf a+\eta J(\mathbf a)\mathbf g$\\
Raw+$R$ & $R(\mathbf a+\eta\mathbf g)$\\
RetraQ & $R(\mathbf a+\eta J(\mathbf a)\mathbf g)$\\
\bottomrule
\end{tabular}
\end{table}

The controlled toy campaign calibrates the median total movement
$\|\mathbf a^\star-\mathbf a\|_2$ to its target displacement.
The deployed calibration described in Section~\ref{sec:method}
instead measures the movement induced by the critic relative to the
mapped baseline,
$\|R(\mathbf a+\eta J(\mathbf a)\mathbf g)-R(\mathbf a)\|_2$.
These quantities differ whenever $R(\mathbf a)\ne\mathbf a$, so the
controlled results evaluate the operator under the toy calibration.
AWR is not displacement-calibrated in the evaluated implementation.
QAM~\citep{qam} fits the actor by adjoint matching, with the frozen BC flow as
the base field; its temperature is matched to the same target displacement by
bisection on the achieved actor drift and reverts to its default value when the
target lies outside the achievable range.

Online component runs can acquire different replay data and critic
parameters as training proceeds, so their differences measure the resulting
learning procedures; the controlled problems of
Section~\ref{sec:exp-geometry} isolate the immediate effect of each operator
at fixed actions and a fixed critic.

\subsection{Forward-Jacobian Diagnostics}
\label{app:retraq-geometry-diagnostics}

Fix a context and a point $\mathbf a$ on its known $k$-dimensional
manifold, with $1\le k<D$.
Let $P_T(\mathbf a)$ and $P_N(\mathbf a)$ be the orthogonal tangent
and normal projectors defined in Section~\ref{sec:prelim}.
We measure
\begin{align}
G_N(\mathbf a)
&:=\frac{\|J(\mathbf a)P_N(\mathbf a)\|_F}{\sqrt{D-k}},
\label{eq:retraq-normal-gain}\\
E_T(\mathbf a)
&:=\frac{\|(J(\mathbf a)-I_D)P_T(\mathbf a)\|_F}{\sqrt{k}},
\label{eq:retraq-tangent-error}
\end{align}
where $\|\cdot\|_F$ is the Frobenius norm, the square root of the sum
of squared matrix entries.
Equivalently, these are root-mean-square responses over an orthonormal
normal or tangent basis. They also equal the root-mean-square responses
to uniformly sampled unit directions in the respective subspaces.
The implementation evaluates the basis responses with forward JVPs
without forming the Jacobian.
For $J=P_T$, both diagnostics vanish.
For $J=I_D$, normal gain is one and tangent error is zero.
Small tangent error alone therefore does not establish normal filtering.
These diagnostics describe average directional responses, rather than
worst-direction operator-norm bounds.

Table~\ref{tab:retraq-jvp-probes} reports the $D=16$ diagnostics.
For sampled or perturbed points away from the manifold, the reference
tangent and normal spaces are evaluated at their known manifold
projections.
Each row summarizes 512 query points for one frozen flow at $t_g=0.9$.

\begin{table*}[t]
\centering
\small
\setlength{\tabcolsep}{4pt}
\caption{Forward-Jacobian diagnostics for one $D=16$ flow. Each cell is a quantile over query points.}
\label{tab:retraq-jvp-probes}
\begin{tabular}{@{}lrrrr@{}}
\toprule
Query set & $G_N$, med. & $G_N$, 95th & $E_T$, med. & $E_T$, 95th\\
\midrule
On manifold & 0.0341 & 0.0446 & 0.0597 & 0.3979\\
Actor samples & 0.0334 & 0.0424 & 0.0470 & 0.3483\\
Perturbed points & 0.0342 & 0.0448 & 0.0604 & 0.3977\\
\bottomrule
\end{tabular}
\end{table*}

For this flow, a separate same-input comparison between $J\mathbf v$
and $J^\top\mathbf v$ reports a relative difference of approximately
$0.9167$, after the campaign's aggregation over points and four probe
directions.
Because the learned Jacobian is far from symmetric, we report
forward-Jacobian diagnostics, which match the JVP used by RetraQ.

\begin{table*}[t]
\centering
\small
\setlength{\tabcolsep}{4pt}
\caption{Separate dimension sweep. $G_N$ is the median normal gain at the evaluation time $t_g$. Training budget is fixed across dimensions. Dashes mark two curved cells whose flows did not pass the fit check.}
\label{tab:retraq-dimension-probes}
\begin{tabular}{@{}rrlrrl@{}}
\toprule
$D$ & $k$ & Geometry & $t_g$ & $G_N$ & Model check\\
\midrule
2 & 1 & flat & 0.9844 & 0.0567 & accepted\\
2 & 1 & curved & 0.9531 & 0.0343 & accepted\\
8 & 2 & flat & 0.9844 & 0.0789 & accepted\\
8 & 2 & curved & 0.9844 & 0.0958 & accepted\\
32 & 2 & flat & 0.9844 & 0.2120 & accepted\\
32 & 2 & curved & --- & --- & not accepted\\
128 & 2 & flat & 0.9844 & 0.8642 & accepted\\
128 & 2 & curved & --- & --- & not accepted\\
512 & 2 & flat & 0.9844 & 0.9659 & accepted\\
512 & 2 & curved & 0.9844 & 0.9661 & accepted\\
\bottomrule
\end{tabular}
\end{table*}

The dimension sweep in Table~\ref{tab:retraq-dimension-probes} uses a
different geometric family from the $D=16$ refinement campaign.
Training budget was held fixed across dimensions, so fitting error also
varies across configurations and the sweep does not isolate a causal effect
of dimension. The $D=16$ problem is not part of this sweep.

\paragraph{Guidance-level sweep.}
Figure~\ref{fig:retraq-geometry}a,b repeats the diagnostics on flat manifolds
with $D\in\{2,8,32,128,512\}$ at five guidance levels
$t_g\in\{0.922,0.938,0.953,0.969,0.984\}$, with one flow per dimension trained
for 20{,}000 steps (512 on-manifold query points; 128 for $D=512$). The deployed
level $t_g=0.9$ lies just below this grid. Across the grid, the tangent error
falls from about $0.07$--$0.09$ to about $0.01$--$0.02$, whereas the normal gain
rises, from $0.27$ to $0.86$ at $D=128$ and from $0.84$ to $0.97$ at $D=512$;
for $D\le8$ it stays below $0.1$. Because the training budget is fixed, the
flow fits the high-dimensional cells worst (normalized held-out loss $0.708$ at
$D=512$ against $0.072$--$0.418$ elsewhere), so the growth with $D$ is
confounded with fitting quality. Curved cells at $D=32$ and $128$, whose
flows did not pass the fit check, are omitted.

\paragraph{Critic exploitation.}
On the $D=16$ problem, the critic's predicted objective and the true objective
of the refined actions are $1.498$ and $-16.794$ for raw ascent, $0.679$ and
$-0.582$ for filter-only, $0.831$ and $0.774$ for Raw$+R$, and $0.838$ and
$0.827$ for RetraQ (Figure~\ref{fig:retraq-geometry}c).

\subsection{Controlled Refinement Results}

\begin{table}[t]
\centering
\small
\setlength{\tabcolsep}{4pt}
\caption{Controlled $D=16$ problem after 30 refinement--distillation
rounds. The objective is averaged over 4096 samples within each actor
seed. The distance statistic is the within-seed 95th percentile of
Euclidean distance to the known manifold. Both columns report medians
over eight actor seeds sharing one reference flow and critic.
The map-only probe did not report this distance percentile.}
\label{tab:retraq-toy}
\begin{tabular}{@{}lrr@{}}
\toprule
Method & True objective $\uparrow$ & Distance, 95th pct.\ $\downarrow$\\
\midrule
BC reference       & 0.3947   & 0.0149\\
Self-distillation  & 0.2656   & 0.0812\\
Map-only           & 0.2859   & ---\\
Raw ascent         & -16.7940 & 1.0160\\
Filter-only        & -0.5822  & 0.4300\\
Raw+$R$            & 0.7744   & 0.0726\\
\textbf{RetraQ}      & 0.8272   & 0.0284\\
\bottomrule
\end{tabular}
\end{table}

Table~\ref{tab:retraq-toy} uses the batch mean of the analytic task
objective before taking the median over actor seeds.
Using the batch median instead changes the filter-only result from
$-0.5822$ to $0.1088$.
Both summarize the same campaign but answer different questions about
its sample distribution. The batch mean is retained as the primary
objective statistic so that the poor lower tail remains represented.
The reported distance percentile describes proximity to the known
manifold. It is not a lower bound or a measure of statistical uncertainty.

We report continuous distances rather than threshold-based
out-of-distribution fractions, whose radius would depend on a reference
control that itself drifts.

\subsection{LIBERO Evaluation and Offline Stability}
\label{app:retraq-libero}

\begin{figure}[t]
\centering
\includegraphics[width=\linewidth]{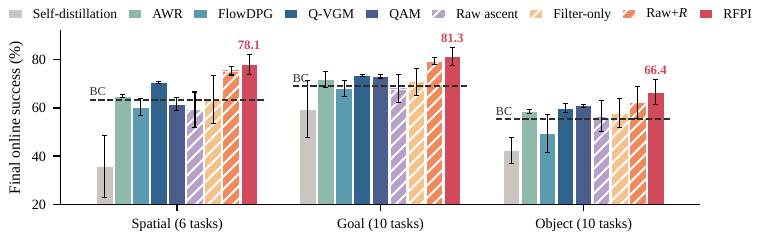}
\caption{\textbf{Final online LIBERO success} (\%), mean $\pm$ standard
deviation over three actor seeds. Cool colors: baselines; hatched warm colors:
component ablations of RetraQ; dashed lines: BC. All numbers are in
Table~\ref{tab:retraq-libero-all}.}
\label{fig:retraq-libero-online}
\end{figure}

\begin{figure}[t]
\centering
\includegraphics[width=\linewidth]{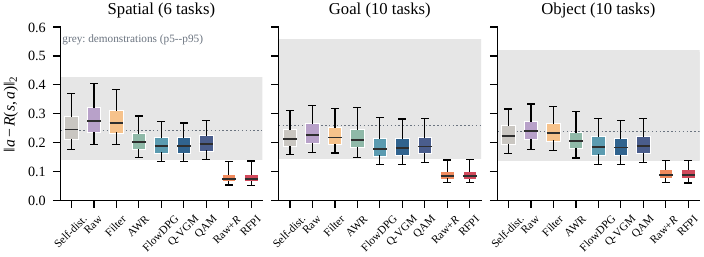}
\caption{\textbf{Retraction residual of the final online actors on LIBERO},
$\|\mathbf a-R(\mathbf s,\mathbf a)\|_2$ at $t_g=0.9$, pooled over tasks and
seeds (boxes: quartiles; whiskers: 5th--95th percentiles). Grey band and dotted
line: the same residual on held-out demonstration chunks. The residual is
computed with the frozen map that RetraQ and Raw$+R$ use to build their targets,
so it measures consistency with the behavior model rather than a distance to
the data.}
\label{fig:retraq-libero-residual}
\end{figure}

Figure~\ref{fig:retraq-libero-residual} shows that the two retracted arms produce
actions near fixed points of the frozen map: more than $99\%$ of their actions
have a smaller residual than the median demonstration chunk, against
$29$--$91\%$ for the other arms. Actors trained without the retraction,
including self-distillation, remain within the range of the demonstrations.

Online LIBERO evaluation uses 30 test initializations per task and actor
seed, indexed 0--29. The reported validation protocol uses five
initializations starting at index 45.
The comparison covers all ten Goal and all ten Object tasks.
The online Spatial evaluation covers six tasks, with task indices
$\{0,1,4,5,6,7\}$. These are the Spatial tasks with the lowest BC success;
on the remaining tasks, BC already leaves little room for improvement through
offline or online fine-tuning.
The offline Spatial campaign covers eight tasks, so the two Spatial
averages are not directly comparable.

\begin{table*}[t]
\centering
\small
\setlength{\tabcolsep}{4pt}
\caption{All online LIBERO arms. Entries are success (\%) as mean $\pm$ across-seed standard deviation. Goal and Object cover all ten tasks. Spatial covers six tasks.}
\label{tab:retraq-libero-all}
\begin{tabular}{@{}lccc@{}}
\toprule
Method & Goal (10 tasks) & Object (10 tasks) & Spatial (6 tasks)\\
\midrule
Self-distillation & $59.56 \pm 11.91$ & $42.22 \pm 5.30$ & $35.74 \pm 12.94$\\
AWR & $71.89 \pm 3.29$ & $58.44 \pm 0.69$ & $64.81 \pm 0.64$\\
FlowDPG & $68.11 \pm 3.20$ & $49.33 \pm 7.97$ & $60.37 \pm 3.61$\\
Q-VGM & $73.56 \pm 0.38$ & $59.89 \pm 1.95$ & $70.56 \pm 0.56$\\
QAM & $73.11 \pm 0.69$ & $60.89 \pm 0.69$ & $61.48 \pm 2.74$\\
Raw ascent & $67.89 \pm 5.82$ & $56.56 \pm 6.38$ & $59.26 \pm 7.34$\\
Filter-only & $70.89 \pm 5.60$ & $57.78 \pm 6.00$ & $63.52 \pm 10.08$\\
Raw+$R$ & $79.44 \pm 1.39$ & $62.11 \pm 6.68$ & $75.37 \pm 1.70$\\
\textbf{RetraQ} & $\mathbf{81.34 \pm 3.71}$ & $\mathbf{66.42 \pm 5.18}$ & $\mathbf{78.10 \pm 4.23}$\\
\bottomrule
\end{tabular}
\end{table*}

For each training seed, task success rates are averaged with equal task
weights. Table~\ref{tab:retraq-libero-all} reports the mean and sample
standard deviation of these seed-level averages.

\begin{table*}[t]
\centering
\small
\setlength{\tabcolsep}{4pt}
\caption{Offline LIBERO success (\%) at each fixed refinement round $r$. Each value averages three actor seeds and 30 evaluation episodes per task and seed, with equal task weights. Shared BC reference rates are given in the text.}
\label{tab:retraq-libero-offline}
\begin{tabular}{@{}lrrrrrrrrr@{}}
\toprule
 & \multicolumn{3}{c}{Spatial (8 tasks)} & \multicolumn{3}{c}{Goal (10 tasks)} & \multicolumn{3}{c}{Object (10 tasks)}\\
\cmidrule(lr){2-4}\cmidrule(lr){5-7}\cmidrule(lr){8-10}
Method & $r=1$ & $r=2$ & $r=3$ & $r=1$ & $r=2$ & $r=3$ & $r=1$ & $r=2$ & $r=3$\\
\midrule
Self-distillation & 64.86 & 61.11 & 62.22 & 66.67 & 66.11 & 66.11 & 50.22 & 51.44 & 51.00\\
AWR & 65.56 & 70.28 & 69.86 & 72.00 & 73.00 & 76.11 & 55.44 & 57.67 & 58.78\\
FlowDPG & 65.00 & 64.31 & 64.58 & 73.67 & 75.44 & 74.33 & 56.67 & 55.89 & 56.44\\
Q-VGM & 19.86 & 23.89 & 21.94 & 34.22 & 35.33 & 32.44 & 19.89 & 18.56 & 17.56\\
QAM & 40.83 & 38.06 & 37.92 & 36.67 & 34.44 & 34.22 & 24.11 & 22.89 & 20.33\\
Raw ascent & 38.89 & 22.08 & 16.67 & 56.11 & 40.33 & 23.33 & 39.22 & 17.44 & 8.89\\
\textbf{RetraQ} & 71.39 & 54.86 & 39.86 & 67.78 & 55.33 & 50.00 & 55.44 & 29.89 & 21.67\\
\bottomrule
\end{tabular}
\end{table*}

\begin{figure}[t]
\centering
\includegraphics[width=\linewidth]{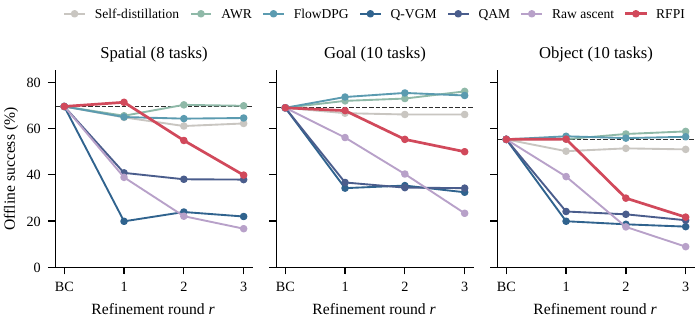}
\caption{\textbf{Repeated offline refinement on LIBERO}
(Table~\ref{tab:retraq-libero-offline}). Round ``BC'' is the shared BC
reference (dashed line).}
\label{fig:retraq-libero-offline}
\end{figure}

\paragraph{Limits of repeated offline improvement.}
\phantomsection\label{sec:exp-boundaries}%
Repeated offline improvement behaves differently from online improvement.
After three refinement--distillation rounds, RetraQ success falls from
the BC reference rates of $69.58\%$, $69.00\%$, and $55.33\%$ to
$39.86\%$, $50.00\%$, and $21.67\%$ on the evaluated Spatial, Goal,
and Object task sets, respectively
(Table~\ref{tab:retraq-libero-offline} and Figure~\ref{fig:retraq-libero-offline}).
AWR and FlowDPG do not exhibit the same across-suite deterioration
under their evaluated configurations.
The self-distillation control also changes over rounds, although its
decline is smaller than RetraQ's.
These results limit claims of reliable repeated offline improvement.
They motivate distinguishing errors in refined targets from errors
introduced when fitting those targets, which the final policy scores
alone cannot separate.

The offline curves in Table~\ref{tab:retraq-libero-offline} evaluate fixed
rounds and show the complete available suite aggregates.
Round is denoted by $r$ to distinguish it from the intrinsic dimension
$k$ used in the geometric analysis.
The shared BC reference is measured once per task.
Raw+$R$ and map-only were not evaluated in this offline LIBERO campaign.
On Spatial, AWR and QAM were evaluated after a simulator reinstallation that
changed the outcomes of re-evaluated earlier cells, so their Spatial offline
entries are not directly comparable with those of the other arms.

The offline BC reference means are $69.58\%$ for the eight Spatial
tasks, $69.00\%$ for Goal, and $55.33\%$ for Object.
Restricting the Spatial BC evaluations to the six online
task indices gives $63.33\%$.
Figure~\ref{fig:retraq-libero-online} uses $63.33\%$, $69.00\%$, and $55.33\%$ as
the BC references for the online Spatial, Goal, and Object suites.

\subsection{RoboMimic Held-Out Comparison}

The preferred Can comparison fixes the number of refinement rounds
at two using evaluation seeds 90000--90039, then evaluates on
held-out seeds 90100--90199.
Each trained arm has three actor-training seeds and 100 held-out
episodes per seed.
The BC reference has one checkpoint evaluated on 100 episodes.
The broader 90000--90199 evaluation band includes selection episodes
and is not used as the primary held-out estimate.

\begin{table}[t]
\centering
\small
\setlength{\tabcolsep}{4pt}
\caption{RoboMimic Can held-out success (\%) after two refinement rounds. The final column is the range across the three actor seeds, not a confidence interval. The BC reference has one checkpoint.}
\label{tab:retraq-can-heldout}
\begin{tabular}{@{}lrr@{}}
\toprule
Method & Mean & Across-seed range\\
\midrule
BC reference & 53.00 & one checkpoint\\
FlowDPG & 64.33 & 62.00--67.00\\
Q-VGM & 69.00 & 68.00--71.00\\
Raw+$R$ & 67.00 & 65.00--69.00\\
\textbf{RetraQ} & 70.33 & 64.00--77.00\\
\bottomrule
\end{tabular}
\end{table}

For RetraQ minus Raw+$R$, the paired 95\% interval is
$[-2.00,8.67]$ percentage points around a difference of $3.33$ points.
It was constructed by resampling episode pairs within each training
seed and is conditional on those trained policies.
The reported McNemar value is $p=0.268$.
The difference is therefore not statistically significant at this sample
size.

\paragraph{Critic quality.}
\phantomsection\label{app:retraq-critic-quality}%
Table~\ref{tab:retraq-critic-quality} repeats the Can refinement with three
critics trained on $588$, $3{,}234$, and $4{,}410$ samples. Each entry is the
best success over refinement rounds $r\ge1$ on the selection band
(seeds 90000--90039, 40 episodes per actor seed), and the round is selected on
the same episodes, so the entries are optimistic peaks rather than held-out
estimates. The BC rate on this band is $57.5\%$. With the weakest critic, RetraQ
does not exceed BC and ranks last; with the two stronger critics, it ranks
first. At $4{,}410$ samples, only Q-VGM has three actor seeds.
For the RetraQ arm, the filtered and raw gradients have mean cosine $0.670$,
$0.697$, and $0.712$ for the three critics, with norm ratios
$\|J_R\mathbf g\|/\|\mathbf g\|$ of $0.400$, $0.342$, and $0.349$ (first round,
averaged over seeds).

\begin{table}[t]
\centering
\small
\setlength{\tabcolsep}{4pt}
\caption{RoboMimic Can success (\%) versus critic training-set size. Best
round per method, selected on the same selection-band episodes; BC is
$57.5\%$. $^\ddagger$One actor seed; all other cells have three.}
\label{tab:retraq-critic-quality}
\begin{tabular}{@{}lrrr@{}}
\toprule
Critic samples & RetraQ & Q-VGM & FlowDPG\\
\midrule
$588$     & $57.5$ & $59.2$ & $62.5$\\
$3{,}234$ & $\mathbf{73.3}$ & $56.7$ & $71.7$\\
$4{,}410$ & $\mathbf{87.5}^\ddagger$ & $79.2$ & $80.0^\ddagger$\\
\bottomrule
\end{tabular}
\end{table}

\subsection{OGBench Online Comparison}

The cube-single campaign starts from an offline-refined actor.
The reported initial success is $30.7\%$ for that actor and $8.0\%$
for the BC reference.
Table~\ref{tab:retraq-ogbench} reports final policies with two training
seeds and 200 evaluation episodes per seed.

\begin{table}[t]
\centering
\small
\setlength{\tabcolsep}{4pt}
\caption{Final OGBench cube-single success (\%), mean $\pm$ standard deviation over two seeds.}
\label{tab:retraq-ogbench}
\begin{tabular}{@{}lc@{}}
\toprule
Method & Success\\
\midrule
Continued BC & $3.50 \pm 1.41$\\
Self-distillation & $19.75 \pm 13.79$\\
Map-only & $0.25 \pm 0.35$\\
Raw ascent & $97.00 \pm 3.54$\\
Raw+$R$ & $99.00 \pm 0.71$\\
\textbf{RetraQ} & $98.75 \pm 0.35$\\
Q-VGM & $95.00 \pm 5.66$\\
FlowDPG & $4.50 \pm 0.71$\\
\bottomrule
\end{tabular}
\end{table}

The table's standard deviations summarize two training seeds.

\paragraph{Filter diagnostics.}
During online training of the RetraQ arm, we record the cosine between the
filtered and raw critic gradients and their norm ratio at 10k, 25k, 50k, and
100k steps for both seeds. The median cosine is $0.988$--$0.991$ at every
checkpoint, and the ratio of median norms $\|J_R\mathbf g\|/\|\mathbf g\|$ is
$0.42$--$0.48$. On this task, the filter therefore changes mainly the length of
the step, which the displacement calibration then sets, consistent with the
near-equal final success of RetraQ and Raw$+R$ in Table~\ref{tab:retraq-ogbench}.

On the harder cube-double task, both evaluated methods remain near zero (RetraQ
$2.50\%$, Q-VGM $1.25\%$; two seeds, 400 episodes each), so we do not draw
conclusions from it.

\subsection{Is RetraQ Behavior Regularization?}
\label{app:retraq-behavior-reg}

A natural concern is that, at $\sigma_g\approx0.11$, the map $R$ might act
mainly as a pull toward the behavior policy and $J_R$ as an isotropic
shrinkage, so that RetraQ would reduce to critic ascent plus behavior
regularization. The evidence collected above bears on this question as follows.

\paragraph{An explicit behavior anchor does not reproduce the retraction.}
On LIBERO, the component ablations share the distillation objective of
Eq.~\eqref{eq:retraq-distillation}, including its BC anchor, and differ only in
their targets (Table~\ref{tab:retraq-operator-definitions}). With this anchor,
raw ascent and filter-only remain within about two points of BC on
average, whereas replacing the target $\mathbf a+\eta\mathbf g$ by
$R(\mathbf a+\eta\mathbf g)$ adds $5.6$--$16.1$ points
(Table~\ref{tab:retraq-libero-all}). The retraction therefore contributes beyond
the behavior regularization that the anchor already provides.

\paragraph{The map alone does not improve the policy.}
Map-only targets $R(\mathbf a)$ apply the same map without a critic step. They
do not improve over the behavior policy: the true objective is $0.286$ against
$0.395$ for BC on the $D=16$ problem (Table~\ref{tab:retraq-toy}), and OGBench
success is $0.25\%$ against $3.50\%$ for continued BC
(Table~\ref{tab:retraq-ogbench}). Self-distillation also falls below BC on
LIBERO. The gains thus require the critic direction; $R$ contributes by
correcting the critic step rather than by regularizing on its own.

\paragraph{RetraQ moves along the manifold, not back to BC.}
On the $D=16$ problem, where the manifold is known, a pull toward the behavior
policy would move refined actions toward BC and their objective toward BC's
$0.395$. RetraQ instead reaches a true objective of $0.827$, while the
95th-percentile distance to the manifold changes only from $0.015$ for BC to
$0.028$, against $0.073$ for Raw$+R$ (Table~\ref{tab:retraq-toy}). The
improvement therefore comes from motion along the manifold. On this problem,
the learned $J_R$ has median normal gain $0.034$ and tangent error $0.060$ on
the manifold (Table~\ref{tab:retraq-jvp-probes}): it is close to $P_T$, not to a
scaled identity.

\subsection{Real-Robot Setup and Records}
\label{app:retraq-real-robot}

\begin{table}[t]
\centering
\small
\setlength{\tabcolsep}{5pt}
\caption{Real-robot phone-charging success rates (\%) without online
interaction, as plotted in Figure~\ref{fig:retraq-real}b.
\textbf{Top:} test-time guidance of the frozen BC policy; the BC reference is
shared across the two critic columns, and all guidance methods are
displacement-matched to RetraQ. Each entry is over 60 trials.
\textbf{Bottom:} offline distillation with the deployed guidance level
$t_g=0.9$ and a larger displacement budget shared by the three arms; the RetraQ
arm uses the Critic~1 fork critic.}
\label{tab:retraq-real}
\begin{tabular}{@{}lcc@{}}
\toprule
Method & Critic 1 & Critic 2\\
\midrule
\multicolumn{3}{@{}l}{\emph{Test-time guidance (policy parameters fixed)}}\\
BC                  & \multicolumn{2}{c}{55.00}\\
QGF~\citep{qgf}      & 58.33 & 56.67\\
Q-VGM               & 0.00  & 58.33\\
FlowDPG             & 63.33 & 66.67\\
\textbf{RetraQ}       & \textbf{76.67} & \textbf{73.33}\\
\midrule
\multicolumn{3}{@{}l}{\emph{Offline distillation (shared displacement budget)}}\\
Q-VGM               & 0.00  & ---\\
FlowDPG             & 28.33 & ---\\
\textbf{RetraQ}       & \textbf{60.00} & ---\\
\bottomrule
\end{tabular}
\end{table}

\begin{table}[t]
\centering
\small
\setlength{\tabcolsep}{6pt}
\caption{Real-robot offline-to-online success (\%) versus online training
time, as plotted in Figure~\ref{fig:retraq-real}c. All methods start from the
same BC policy and the same warmed-up IQL critic (0\,h); each then trains its
own critic on its own rollouts under the same total interaction budget.
Each checkpoint of each training run is evaluated with 60 trials, and each
entry is the mean over three training runs per method.}
\label{tab:retraq-real-online}
\begin{tabular}{@{}lccc@{}}
\toprule
Method & 0\,h & 0.5\,h & 1.0\,h\\
\midrule
QGF     & 55.00 & 88.33 & 91.67\\
FlowDPG & 55.00 & 88.33 & 86.67\\
Q-VGM   & 55.00 & 95.00 & 93.33\\
\textbf{RetraQ} & 55.00 & \textbf{98.33} & \textbf{100.00}\\
\bottomrule
\end{tabular}
\end{table}

\paragraph{Platform and policy.}
The task is phone-charging plug insertion on an AgiBot G02 dual-arm robot
controlled at 30\,Hz, with three cameras (head and both wrists). The policy
is a $\pi_{0.5}$ flow behavior-cloned on task demonstrations for 80{,}000
steps (batch size 256). It predicts 50-frame chunks, of which the first 30
frames (1\,s) are executed, and its flow is integrated with 10 steps.

\paragraph{Test-time guidance.}
RetraQ uses $t_g=0.9$ ($\sigma_g\approx0.11$), $\eta=0.01$ calibrated once
offline and then frozen, $d_{\mathrm{target}}=0.025$, and a displacement cap
of $0.25$ (Appendix~\ref{app:retraq-update-details}). Guidance is enabled only
at detected fork states~0--2 (Appendix~\ref{app:retraq-policy}); elsewhere the
BC chunk is executed unchanged. The baselines use the same frozen policy,
the same critic, and the same gate: QGF updates the velocity as
$v\gets v-\lambda\mathbf g$ without a $1/t$ factor, FlowDPG takes a normalized
velocity step with a $1/t$ factor, and Q-VGM follows its first
version~\citep{qvgm_v1}: it takes gradient-clipped ascent steps along
$\mathbf g$ from the one-step clean-action estimate, clips each iterate to the
action range, and keeps the highest-valued iterate, with gating and a $1/t$
factor. Their step sizes are set by
bisection so that their median guided displacement equals the median
displacement of the deployed RetraQ arm.

\paragraph{Critics and trials.}
Every real-robot evaluation consists of 60 trials. In offline-to-online
learning, this holds for every checkpoint of each of the three training runs
per method, so each reported online entry averages $3\times60$ trials.
Critic~1 and Critic~2 are the linear and the nonlinear fork critic
of Appendix~\ref{app:retraq-data}, trained on the same data.

\paragraph{Why Q-VGM fails with the linear critic.}
With Critic~1, $A_\omega$ in Eq.~\eqref{eq:fork-critic} is linear in the
action deviation, so $\nabla_{\mathbf a}Q$ points in the same direction at
every action, and the critic value increases monotonically along it. Q-VGM
takes several ascent steps along this direction, clips each iterate to the
action range, and keeps the highest-valued iterate. Under a monotone critic,
this rule always selects the last, farthest iterate, and neither the
coordinate clip nor the value-based selection removes the normal component of
the fixed direction. The executed chunks can therefore leave the region
covered by the behavior data at the fork states, which is consistent with the
observed failure of every trial. Because the step size is matched to RetraQ's median displacement,
this failure stems from the direction of the update rather than its size.
With the nonlinear Critic~2, the gradient direction varies with the action and
the value need not increase monotonically along the ascent path, and Q-VGM
stays near BC ($58.33\%$). RetraQ receives the same
gradient but removes its normal component with $J_R$ and maps the endpoint
back with $R$.

\paragraph{Offline distillation.}
The offline-distillation arms are a separate experiment. They distill
refined targets into the policy without online interaction, under one shared
displacement budget: the total RetraQ displacement ($2.10$ at $\eta=0.05$)
measured at this setting. The RetraQ arm uses the Critic~1 fork critic and the
same guidance level as deployment, $t_g=0.9$ ($t'_g=0.1$ in the $\pi_{0.5}$
convention). The displacement budget, however, is much larger than that of
test-time guidance ($d_{\mathrm{target}}=0.025$), far outside the perturbation range that supervises the fork
critic, so it probes how each method's targets behave when the critic is
queried away from its training data. The Q-VGM-distilled policy reaches
$0\%$: it repeatedly hits the table. The FlowDPG-distilled policy reaches $28.33\%$: it approaches the
socket but frequently misaligns the plug at insertion. The RetraQ-distilled
policy reaches $60.00\%$, on par with the BC rate of $55\%$ (Fisher's exact
test, $p=0.71$). At this budget,
unconstrained targets thus degrade the distilled policy, whereas RetraQ targets
preserve it; a comparison that varies only $d_{\mathrm{target}}$ would be needed
to determine whether smaller displacements also let offline RetraQ improve
substantially over BC.

\begin{figure}[t]
\centering
\includegraphics[width=\linewidth]{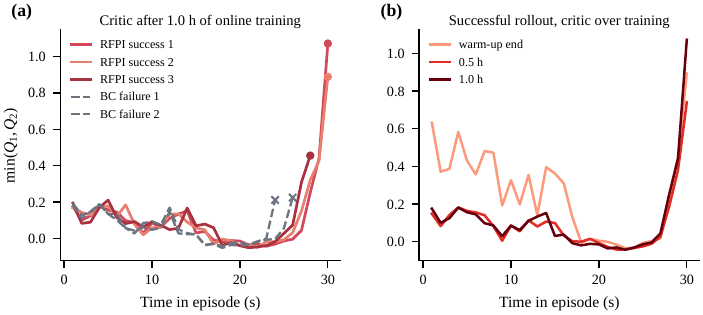}
\caption{\textbf{Online IQL critic along fixed evaluation rollouts}
($\min(Q_1,Q_2)$ at the end of each 1\,s action chunk).
\textbf{(a)}~Critic after 1.0\,h of online training along three
successful rollouts of the online RetraQ policy and two failed rollouts of the
BC policy; markers show the last chunk. The value rises sharply only in the
final seconds of successful insertions. The two groups come from different
policies, so this panel is illustrative.
\textbf{(b)}~One successful rollout scored by the critic at successive
checkpoints. After warm-up, the critic assigns high values early in the
episode; with online training, it concentrates value at the insertion.}
\label{fig:retraq-robot-critic}
\end{figure}

\paragraph{Online RetraQ.}
Averaged over three training runs, online RetraQ reaches $98.33\%$ after
30 minutes and $100\%$ after one hour of online training, the highest success
of all methods at each checkpoint (Table~\ref{tab:retraq-real-online}). It uses the IQL critic and the
update schedule of Appendix~\ref{app:retraq-policy}: every 20 new chunk
transitions allow one actor update attempt and two critic updates, with
batches drawn entirely from online replay. No human interventions occurred
during training; episodes are labeled by the operator as successes or
failures.

\end{document}